\pdfoutput=1

\documentclass[preprint,12pt]{elsarticle}

\usepackage{amssymb}
\usepackage{multirow}
\usepackage{color}
\usepackage{url}
\usepackage{amsmath}
\usepackage{hyperref}

\usepackage{array}
\newcolumntype{M}[1]{>{\centering\arraybackslash}m{#1}}

\newcommand{\revise}[1]{{\color{black} #1}}

\newcommand{\stitle}[1]{\vspace{1ex}\noindent{\ttfamily{\em #1}}}

\newcommand{\eg}{\emph{e.g.}}
\newcommand{\ie}{\emph{i.e.}}

\newcommand{\etal}{\emph{et al.}}

\usepackage[normalem]{ulem}
\definecolor{gray}{rgb}{0.55,0.55,0.55}

\journal{Visual Informatics}

\begin{document}

\begin{frontmatter}



\title{Generative AI for Visualization: State of the Art and Future Directions}


\author[inst1,inst2]{Yilin Ye}
\author[inst1]{Jianing Hao}
\author[inst1]{Yihan Hou}
\author[inst1]{Zhan Wang}
\author[inst1]{Shishi Xiao}
\author[inst1,inst2]{Yuyu Luo}
\author[inst1,inst2]{Wei Zeng}

\affiliation[inst1]{organization={The Hong Kong University of Science and Technology (Guangzhou)},
            city={Guangzhou},
            state={Guangdong},
            country={China}}

\affiliation[inst2]{organization={The Hong Kong University of Science and Technology},
            state={Hong Kong SAR},
            country={China}}




\begin{abstract}

Generative AI (GenAI) has witnessed remarkable progress in recent years and demonstrated impressive performance in various generation tasks in different domains such as computer vision and computational design.
Many researchers have attempted to integrate GenAI into visualization framework, leveraging the superior generative capacity for different operations.
Concurrently, recent major breakthroughs in GenAI like diffusion model and large language model have also drastically increase the potential of GenAI4VIS.
From a technical perspective, this paper looks back on previous visualization studies leveraging GenAI and discusses the challenges and opportunities for future research.
Specifically, we cover the applications of different types of GenAI methods including sequence, tabular, spatial and graph generation techniques for different tasks of visualization which we summarize into four major stages: data enhancement, visual mapping generation, stylization and interaction.
For each specific visualization sub-task, we illustrate the typical data and concrete GenAI algorithms, aiming to provide in-depth understanding of the state-of-the-art GenAI4VIS techniques and their limitations.
Furthermore, based on the survey, we discuss three major aspects of challenges and research opportunities including evaluation, dataset, and the gap between end-to-end GenAI and generative algorithms.
By summarizing different generation algorithms, their current applications and limitations, this paper endeavors to provide useful insights for future GenAI4VIS research.

\end{abstract}




\end{frontmatter}



\section{Introduction}
\label{sec:intro}

VizDeck~\cite{alicia_2012_vizdeck}.
Visualization is a process of rendering graphical representations of spatial or abstract data to assist exploratory data analysis.
Recently, many researchers have attempted to apply artificial intelligence (AI) for visualization tasks~\cite{zhu2020survey, wu2021ai4vis, wang2021survey, wang2022dl4scivis, DBLP:journals/vldb/QinLTL20}.
Particularly, as visualization essentially involves representations and interactions for raw data, many visualization researchers have started to adopt the rapidly developing generative AI (GenAI) technology, a type of AI technology that empowers the generation of synthetic content and data by learning from existing man-made samples~\cite{Abukmeil_2021_survey, Gui_2023_gan}. 
GenAI has come to the foreground of artificial intelligence in recent years, with profound and widespread impact on various research and application domains such as artifact and interaction design (\eg~\cite{hou2024c2ideas,xiao2024typedance,huang2024plantography}).

Recently, multi-modal AI generation model such as Stable Diffusion~\cite{rombach2022high} or DaLL-E 2~\cite{ramesh2022hierarchical} enable laymen users without traditional art and design skills to easily produce high-quality digital paintings or designs with simple text prompts.
In natural language generation, large language models like GPT~\cite{openai2023gpt4} and LLaMa~\cite{touvron2023llama} also demonstrate astounding power of conversation, reasoning and knowledge embedding.
In computer graphics, recent models like DreamFusion~\cite{poole2022dreamfusion} also shows impressive potential in 3D generation.
GenAI's unique strength lies in its flexible capacity to model data and generate designs based on implicitly embedded knowledge gleaned from real-world data.
This characteristic positions GenAI as a transformative force capable of alleviating the workload and complexity associated with traditional computational methods, and extending the diversity of design with more creative generated results than previous methods.

The burgeoning potential of GenAI is particularly evident in its ability to enhance and streamline operations throughout the data visualization process. From data processing to the mapping stage and beyond, GenAI can play a pivotal role in tasks such as data inference and augmentation, automatic visualization generation, and chart question answering.
For instance, the automatic visualization generation has been a longstanding research focus predating the current wave of GenAI methods, offering non-expert users an efficient means of conducting data analysis and crafting visual representations (\eg,~\cite{luo2018deepeye, cui2019text}).
Traditionally, automatic visualization approaches relied on expert-designed rules rooted in design principles~\cite{mackinlay2007show}.
However, these methods were shackled by the constraints of knowledge-based systems~\cite{chen2008data}, struggling to comprehensively incorporate expert knowledge within convoluted rules or oversimplified objective functions.
The advent of GenAI introduces a paradigm shift, promising not only increased efficiency but also a more intuitive and accessible approach to visualization in an era marked by unprecedented technological advancements.

Despite the impressive capability of GenAI, when applied to visualization it can face many challenges because of its unique data structure and analytic requirements.
For example, the generation of visualization images is significantly different from generation of natural or artistic images.
First, the evaluation of GenAI for visualization tasks is more complex than natural image generation as many factors beyond image similarity need to be considered, such as efficiency~\cite{tufte2001visual} and data integrity~\cite{xiao2023let}.
Second, compared to general GenAI tasks trained on large datasets with simple annotations, the diversity and complexity of visualization tasks demand more complex training data~\cite{chen2023state}, which is harder to curate.
Third, the gap between the traditional visualization pipeline with strong rule-based constraints makes it difficult to fully integrate with end-to-end GenAI methods. 
These unique characteristics makes it less straightforward to leverage the latest pre-trained GenAI models in general domain to empower visualization-specific generation.
Therefore, it is important to understand how previous works have utilized GenAI for various visualization applications, what challenges are met and especially how the GenAI methods are adapted to the tasks.

Although some previous surveys have covered the use of AI in a general sense for visualization~\cite{wu2021ai4vis}, to the best of our knowledge, no study has focused on comprehensive review of GenAI methods used in visualization.
This survey extensively reviews the literature and summarizes the AI-powered generation methods developed for visualization.
We categorize the various GenAI methods according to the concrete tasks they address, which correspond to different stages of visualization generation.
In this way, we manage to collect \revise{81} research papers on GenAI4VIS.
We particularly focus on the different algorithms used in specific tasks in the hope of helping researchers understand the state-of-the-art technical development as well as challenges.
We also discuss and highlight potential research opportunities.

This paper is structured as follows.
Section~\ref{sec:scope} outlines the scope and taxonomy of our survey with definition of key concepts.
Starting from Section~\ref{sec:data} to Section~\ref{sec:interact}, each section corresponds to a stage in the visualization pipeline where GenAI has been used.
Specifically, Section~\ref{sec:data} concerns the use of GenAI for data enhancement.
Section~\ref{sec:vis_map} summarizes works leveraging GenAI for visual mapping generation.
Section~\ref{sec:style} focuses on how GenAI is utilized for stylization and communication with visualization.
Section~\ref{sec:interact} covers GenAI techniques to support user interaction.
Each subsection in Section~\ref{sec:data} to Section~\ref{sec:interact} covers a specific task in the stage.
Instead of listing the works one by one, the structure of the subsection is divided into two parts: data \& algorithm and discussion, for a comprehensive understanding of how the current GenAI method works for data of certain structures and what remains challenging for GenAI in particular tasks.
Finally, Section~\ref{sec:challenge} discusses some dominant challenges and research opportunities for future research.

\section{Scope and Taxonomy}
\label{sec:scope}

\subsection{Scope and Definition}
Generative AI (GenAI) is a type of AI technique that generates synthetic artifacts by analyzing training examples; learning their patterns and distribution; and then creating realistic facsimiles. 
GenAI uses generative modeling and advances in deep learning (DL) to produce diverse content at scale by utilizing existing media such as text, graphics, audio, and video~\cite{Abukmeil_2021_survey, Gui_2023_gan}.
A key feature of GenAI is that it generates new content by learning from data instead of explicit programs.

\noindent
\textbf{GenAI methods categorization.}
Despite the differences between different domain targets of generation ranging from text, code, multi-media to 3D generation, the particular algorithms of generation actually depend on the data structures which show common characteristics across different domains.
\revise{Particularly, in GenAI4VIS applications, categorization based on data structures can facilitate more concrete understanding of the algorithms in relation to the different types of data involved in different visualization tasks.}
Here, we provide an overview of \revise{different types of GenAI in terms of} typical data structures associated with data visualization.

\begin{itemize}
\item
\textbf{Sequence Generation}: This category includes the generation of ordered data, such as text, code, music, videos, and time-series data. Sequence generation models, like LSTMs and Transformers, can be used to create content with a sequential or temporal structure.
\item
\textbf{Tabular Generation}: This category covers the generation of structured data in the form of rows and columns, such as spreadsheets or database tables. Applications include data augmentation, anonymization, and data imputation.
\item
\textbf{Graph Generation}: This category involves generating graph and network structures, such as social networks, molecular structures, or recommendation systems. Models like Graph Neural Networks (GNNs) and Graph Convolutional Networks (GCNs) can be used to generate or manipulate graph-structured data.
\item
\textbf{Spatial Generation}: This category encompasses the generation of both 2D images and 3D models. These data have the common characteristics of spatial data in 3D or 2D projection in Euclidean spaces, which can be represented as pixels, voxels or points with 2D/3D coordinates. 2D generation includes image synthesis, style transfer, and digital art, while 3D generation covers computer graphics, virtual reality, and 3D printing. Techniques like GANs, VAEs, and \revise{PointNet~\cite{qi2017pointnet}} can be used for creating 2D and 3D content.
\end{itemize}

\begin{figure*}[t]
\centering
\includegraphics[width=0.995\textwidth]{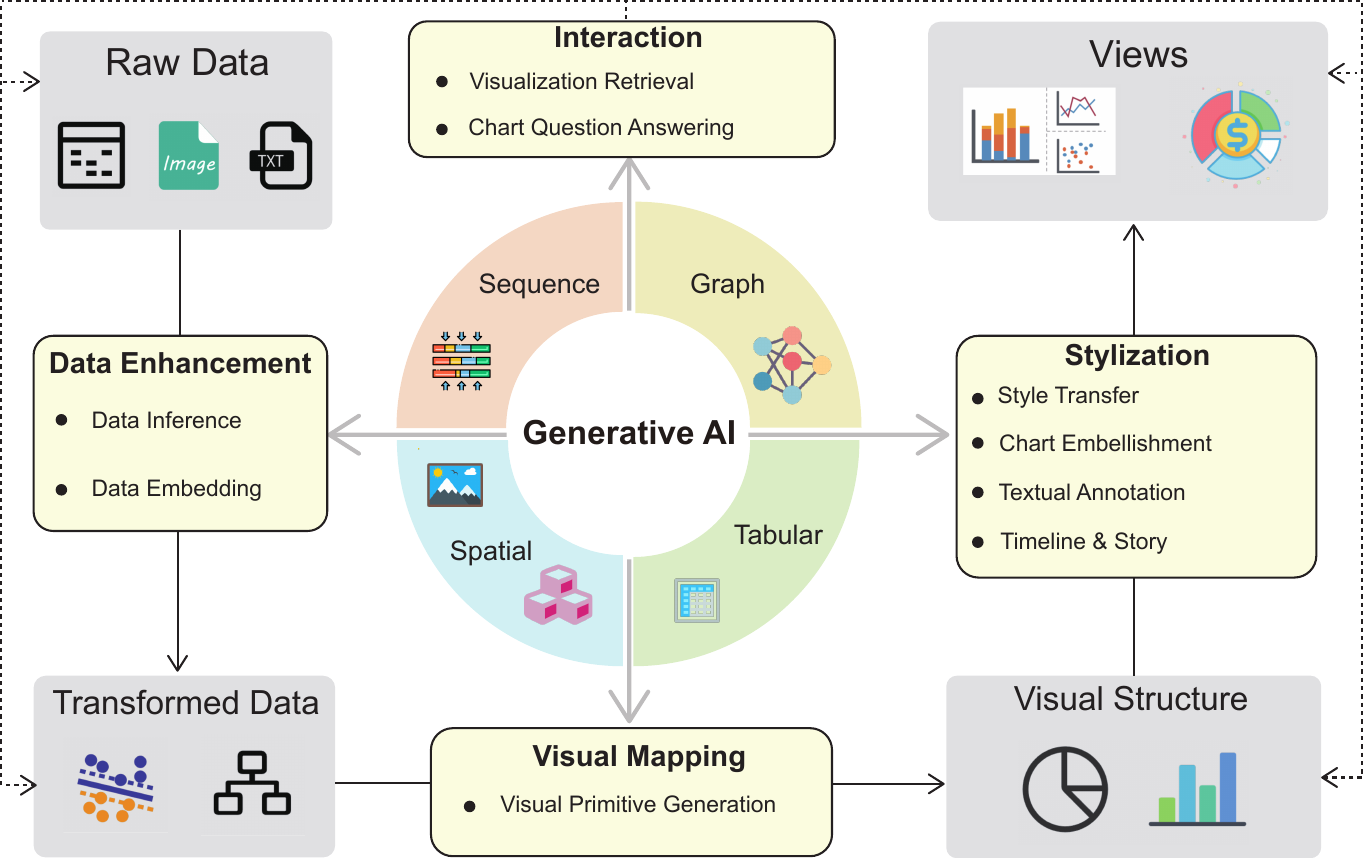}
\vspace{-2mm}
\caption{The overview of GenAI4VIS applications for different visualization tasks, including data enhancement, visual mapping generation, stylization and interaction tasks.}
\label{fig:overview}
\end{figure*}

\noindent
\textbf{GenAI4VIS tasks categorization.}
To categorize and organize the collected articles, we are inspired by the classical visualization pipeline describing different essential stages~\cite{card1999readings}.
However, as GenAI is utilized in broader scenarios different from traditional operations, we also modify the pipeline to encompass some latest research topics.
including \textbf{data enhancement}, \textbf{visual mapping generation}, \textbf{stylization}, and \textbf{interaction}.
Notably, the data transformation part is generalized to the concept of \textbf{data enhancement} inspired by the terminology in the study by McNabb et al.~\cite{mcnabb2017survey}.
In addition, as few GenAI for visualization works focus on the basic view transformation, we replace this part with a broader concept of \textbf{stylization \& communication}.
Under different stages we further categorize the works into specific tasks, as shown in Figure~\ref{fig:overview}.

\begin{itemize}
    \item 
\textbf{Data enhancement}.
Data enhancement refers to the process of improving the quality or completeness of the data or enhancing the feature representation of the data for subsequent visualization.
This can involve 
data augmentation, embedding or other transformations to make it more suitable for visualization.

    \item 
\textbf{Visual mapping generation}. 
This refers to the use of algorithms and software tools to generate visualizations automatically without extensive manual intervention. 
Automatic visual mapping generation allows users to leverage knowledge about how to create appropriate visualization as common wisdom to reduce the workload and man-made violation of design principles.

    \item 
\textbf{Stylization}.
Extending the concept of presentation in~\cite{shen2022towards}, we define stylization in visualization, which involves the application of design principles and aesthetic choices to make the visualization more engaging and effective in conveying information.
It includes decisions about color schemes, fonts, layout, and other visual or textual elements to enhance the information-assisted visualization~\cite{chen2008data}.

\item 
\textbf{Interaction}. 
In the context of data visualization, interaction refers to the dynamic engagement and communication between users and the visualized data. It involves the ability of users to manipulate, explore, and interpret visual representations.
This can involve various forms of interactivity, such as graphical interactions like zooming, panning, clicking and natural language interaction like chart question answering.

\end{itemize}
Earlier methods for these tasks focus on rule-based algorithms with complex expert-designed rules reflecting design principles, which is still effective in many applications such as colormap generation~\cite{tennekes2014tree}.
Some studies also leverage optimization-based methods to minimize expert-defined explicit objective functions.
However, these types of methods differ from GenAI methods in that they are top-down and do not learn from real-world data.
To narrow down the scope of our survey, we exclude all previous generative algorithms that are purely based on rules or optimization.

\begin{figure*}[t]
\centering
\includegraphics[width=0.995\textwidth]{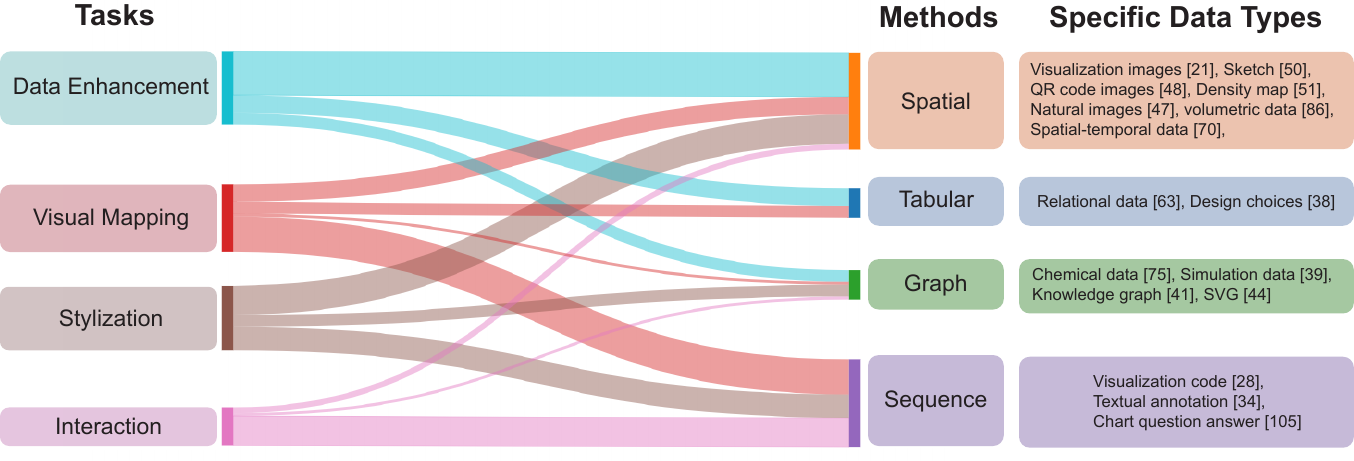}
\vspace{-2mm}
\caption{\revise{Relation between tasks and methods and examples of specific data types involved in different methods.}}
\label{fig:data_type}
\end{figure*}

\revise{
\noindent
\textbf{Relation between different GenAI methods and tasks.}
Due to the wide range of diverse applications in GenAI4VIS, there is no clear-cut one-to-one relation between the type of GenAI methods and the tasks.
Nevertheless, we can observe some interesting correlation.
First, sequence generation is mostly applied in visual mapping or interaction-related tasks.
This is because GenAI such as translation models and the latest LLMs or vision-language model are useful in generating sequence of code specifying visual mapping or sequence of interaction flow and output.
Second, tabular generation is mostly used in data enhancement.
This is because tabular data with attribute columns are the most common initial input data to visualization, which benefits from data enhancement like surrogate data generation for subsequent tasks.
Next, graph generation is also mostly used in data enhancement because data inference and augmentation can facilitate subsequent analysis of graph data. 
However, despite its relatively rare use, it holds great potential for visual mapping and stylization because graphical structure such as knowledge graph or scene graph can benefit optimization of visual encodings and layout.
Finally, spatial generation is mostly applied in data enhancement and stylization tasks.
This is because 2D and 3D data such as images and volumetric data are also common types of input for VIS4AI and SciVis applications, while the embellishment of basic charts into stylized charts relies on image-based generation methods. 
Figure~\ref{fig:data_type} illustrates the relation between GenAI4VIS tasks and methods with a sankey diagram and exemplifies the specific data types that are involved in different methods.  
Table~\ref{tab:method_data} further list the detailed methodologies for each data structure and task.
}

\begin{table}[t]
\centering
\scriptsize
\caption{\revise{Examples of Specific GenAI4VIS methods applied to different tasks and data types.}}
\begin{tabular}{|M{1.3cm}|M{2.8cm}|M{2.6cm}|M{2.5cm}|M{2.5cm}|}\hline
\textbf{} & \textbf{Data Enhancement} & \textbf{Visual Mapping} & \textbf{Stylization} & \textbf{Interaction}\\
\hline
\textbf{Sequence} & - & RNN~\cite{dibia2019data2vis}, Deep Q Network~\cite{zhou2021table2charts}, Transformer~\cite{ncnet}, LLM~\cite{dibia2023lida} & LLM~\cite{liew2022using}, RL~\cite{shi2020calliope}, Detection Network+Template~\cite{liu2020autocaption} & Detection-based Models~\cite{singh2020stl}, Vision-Language Models~\cite{han2023chartllama}\\
\hline
\textbf{Tabular} & Table-GAN~\cite{park2018data} & FFN~\cite{hu2019vizml}, Enumeration+Scoring Network~\cite{luo2018deepeye}  & - & -\\
\hline
\textbf{Graph}  & GNN~\cite{shi2022gnn}, Latent Traversal~\cite{zhang2023graph} & KG embedding~\cite{li2021kg4vis}  & Graph Latent~\cite{kwon2019deep}, Graph Style Extraction Network~\cite{song2022vividgraph} &Graph Contrastive Learning~\cite{li2022structure}\\
\hline
\textbf{Spatial} & VAE~\cite{liu2019latent}, GAN~\cite{han2021stnet}, DRL~\cite{gou2020vatld}, BASNet~\cite{zhang2020viscode}, ISN~\cite{ye_2023_invVis} & Faster R-CNN~\cite{ma2020ladv}, GAN~\cite{chen2019generativemap} & Color Extraction Network~\cite{yuan2021deep}, Siamese Network~\cite{shi2022supporting}, Diffusion~\cite{xiao2023let}, RL~\cite{tang2020plotthread} & Triplet Autoencoder~\cite{luo2023line}, Contrastive Learning~\cite{xiao2023wytiwyr}\\
\hline

\end{tabular}
\label{tab:method_data}
\end{table}

\subsection{Related Survey}
\label{ssec:related_survey}
Some previous surveys cover the applications of artificial intelligence or machine learning in general to information visualization or scientific visualization~\cite{zhu2020survey, wu2021ai4vis, xia2021survey, wang2021survey, wang2022dl4scivis}.
Wu et al.~\cite{wu2021ai4vis} surveyed the development of artificial intelligence technologies applied to information visualization, focusing on three major aspects including visualization data and representation (what), goals (why) and specific tasks (how) which concern the use of AI. 
Other previous surveys touch upon the use of AI in more specific sub-areas of visualization research, such as natural language interface and data story telling~\cite{shen2022towards, chen2023does, he2024leveraging}.
Shen et al.~\cite{shen2022towards} summarizes all the existing technologies supporting natural language interface to different stages of data visualization, including both traditional rule-based techniques and recent AI-powered methods.
Bartolomeo et al.~\cite{di2023doom} envisions the potential of GenAI applied to different stages of visualization, mostly focused on interviewing experts and discussing usage scenarios.
Another closely relevant survey~\cite{yang2023foundation} recently focuses on the two-way relationship between visualization and foundation AI models, which include some large scale GenAI models like GPT.  
In comparison, the focus of our survey is on the concrete technical advancements and challenges of GenAI models for visualization applications.

No previous survey has been dedicated to the various types of GenAI methods in the context of visualization tasks.
Our survey aims to provide a specialized and comprehensive overview of the GenAI techniques that have been used to generate various data or intermediate representations that are useful for visualization.
We also outline challenges and opportunities for future research on GenAI for VIS.

\begin{table}[t]
\centering
\scriptsize 
\caption{Survey taxonomy and example papers. We classify the collected GenAI4VIS papers into different sub-tasks along visualization pipeline and further classify different GenAI methods into sequence, tabular, spatial and graph generation.}
\begin{tabular}{|M{2.2cm}|M{2.7cm}|M{3cm}|M{4.2cm}|}
\hline
\textbf{Tasks} & \textbf{Subtasks} & \textbf{Description} &  
\multicolumn{1}{c|}{\textbf{Examples}} \\
\hline
\multirow{2}{*}{Data Enhance} & Data Inference & Increase samples or dimensions
& tabular~\cite{fan13relational}~\cite{chen2019faketables}~\cite{park2018data} ~\cite{qinl2022synthesizing}~\cite{xu2019modeling}~\cite{zhang2017privbayes}, spatial \cite{liu2019latent}~\cite{wang2019deeporgannet}~\cite{gou2020vatld}~\cite{he2021can}~\cite{zhou2017volume}~\cite{wiewel2019latent} ~\cite{han2021stnet}~\cite{wang2023drava}~\cite{wang2020scanviz}~\cite{he2021can}~\cite{gou2020vatld}~\cite{evirgen2023ganravel}, graph~\cite{singh2020chemoverse}~\cite{zhang2023graph}~\cite{zheng2023desirable}~\cite{shi2022gnn}
\\
\cline{2-4}
& Data Embedding & Embed data to hide information
& spatial~\cite{ye_2023_invVis} ~\cite{zhang2020viscode} ~\cite{fu2020chartem}
\\
\hline

Visual Mapping & Visual Primitive & Generate basic visual structures
& sequence~\cite{dibia2019data2vis} ~\cite{zhou2021table2charts} ~\cite{ADVISor} ~\cite{ncnet} ~\cite{nvbench} ~\cite{song2022rgvisnet} ~\cite{dibia2023lida} ~\cite{wang2023llm4vis} ~\cite{shi2023nl2color} ~\cite{li2024prompt4vis} ~\cite{tian2024chartgpt} ~\cite{li2024visualization}, tabular~\cite{hu2019vizml} ~\cite{luo2018deepeye} ~\cite{luo2018deep} ~\cite{cui2019text}, spatial ~\cite{ma2020ladv} ~\cite{chen2019generativemap} ~\cite{berger2018generative} ~\cite{hong2019dnn} ~\cite{wu2022diver} ~\cite{gan2023v4d}, graph~\cite{li2021kg4vis}
\\
\hline
\multirow{4}{*}{Stylization} & Style Transfer & Imitate styles of examples
& spatial~\cite{yuan2021deep} ~\cite{huang2021visual} ~\cite{liu2022image} ~\cite{chen2019towards} ~\cite{shi2022supporting} ~\cite{lu2020exploring}, graph~\cite{kwon2019deep} ~\cite{wang2019deepdrawing} ~\cite{song2022vividgraph} 
\\
\cline{2-4}
& Embellishment & Generate infographics
& spatial~\cite{dibia2023lida} ~\cite{xiao2023let} ~\cite{wu2023viz2viz}
\\
\cline{2-4}
& Text Annotation & Add information with text
& sequence~\cite{liu2020autocaption, liew2022using, ko2023natural}
\\
\cline{2-4}
& Timeline \& Story & Generate data story
& sequence~\cite{shen2023data} ~\cite{shi2023support} ~\cite{ying2023reviving} ~\cite{wu2023socrates} ~\cite{shi2020calliope}, spatial~\cite{tang2020plotthread}, graph~\cite{ying2023reviving}
\\
\hline

\multirow{2}{*}{Interact} & Retrieval & Find similar charts
& spatial~\cite{xiao2023wytiwyr} ~\cite{luo2023line}, graph~\cite{li2022structure}
\\
\cline{2-4}
& CQA & Answer questions about chart
& sequence~\cite{methani2020plotqa} ~\cite{reddy2019figurenet}, ~\cite{kafle2018dvqa} ~\cite{singh2020stl} ~\cite{chaudhry2020leaf} ~\cite{masry2022chartqa} ~\cite{kafle2020answering} ~\cite{masry2023unichart} ~\cite{han2023chartllama} ~\cite{zhao2024leva}
\\
\hline

\end{tabular}
\label{tab:taxonomy}
\end{table}
\subsection{Survey Methodology}
We combine search-based and reference-driven methods to discover relevant literature.
We first collect relevant papers from previous surveys and recent works.
Then we expand the list by going through the papers' references and citations.
In this process, we also supplement the results by searching with key words in the titles of previously collected papers in both the ACM and IEEE libraries as well as arxiv.
In the paper selection process, we manually filter out the non-GenAI traditional methods such as purely rule-based or optimization-based methods, stressing the key characteristics of GenAI which has learned from real data in self-supervised pre-training or supervised training stages. 
In total we collect \revise{81 papers as listed in Table~\ref{tab:taxonomy}} utilizing GenAI for visualization tasks.
As shown in Table~\ref{tab:taxonomy}, different types of GenAI4VIS techniques include sequence generation, tabular generation, spatial generation and graph generation, which can benefit visualization tasks in different stages such as data enhancement, visual mapping, stylization and interaction. 
Sequence and spatial generation are more often used as they concern the general visualization code, images and natural language interaction.
We acknowledge that our search method may not be exhaustive due to the manually collection through keyword search and citation traversal.
Therefore, this survey mainly provides a comprehensive overview of state-of-the-art GenAI4VIS methods, where application papers with similar methods may not be enumerated.

\section{Data Enhancement}
\label{sec:data}

\subsection{Data Inference}
GenAI can be useful for inferring unobserved data items or data features based on the distributions and feature values of existing data, such as data interpolation, data augmentation and super-resolution, \revise{which we use a general term data inference to describe, as these tasks all aim at inferring unseen data.}

\subsubsection{Graph Generation}

\stitle{Data}. GenAI for graph data inference is commonly applied in the domain of chemical data.

\begin{itemize}
\item
\emph{Chemical data}.
GenAI-powered data interpolation has been used for interactive exploration of chemical data~\cite{gomez2018automatic, singh2020chemoverse} to assist discovery of new molecule structures.
For example, ChemoVerse~\cite{singh2020chemoverse} is an interactive system that leverages interpolation powered by GenAI to help experts understand AI drug design models and verify potential new designs.
\item 
\revise{
\emph{Graph simulation data}.
Graph generation can also be applied to inference of certain physical simulation data which can be modeled as graph data structure, such as ocean simulation data~\cite{shi2022gnn}.
}

\end{itemize}

\stitle{Method}. Typical methods include GNN and latent space traversal:

\begin{itemize}
\item 
\emph{Graph Neural Network (GNN)}. GNN has been developed to model data that can be represented as graphs~\cite{wu2020comprehensive}.
By extracting graph features with operations like graph convolution, GNN can be applied to a wide range of non-Euclidean data with complex relationships, which can also benefit some visualization tasks such as structure-aware visualization retrieval~\cite{li2022structure} and data reconstruction~\cite{shi2022gnn}.
For example, GNN-Surrogate~\cite{shi2022gnn} is proposed to reconstruct ocean simulation data based on simulation parameters for efficient parameter space exploration.
Particularly, because training an end-to-end model that directly reconstruct the full high-resolution ocean simulation data is expensive, the authors proposed to construct an intermediate graph representation for adaptive resolution.
The hierarchical graphs are constructed with a series of operations including edge-weighted graph construction, graph hierarchy generation and hierarchical tree cutting.
GNN-Surrogate, which is an up-sampling graph generator, first transforms input parameters into latent vector.
Then the latent vector is reshaped into initial graph, which is passed through multiple steps of graph convolutions with residual connections.
Specifically, graph convolution generates the features of each node by weighted sum of features in the previous layer for the node and all its neighbors.
In chemical data interpolation, sometimes special latent space traversal algorithm need to be developed to generate desirable intermediate samples, because the direct linear interpolation assumes that the latent space is flat and Euclidean~\cite{white2016sampling}, which may poorly model the complex structure, particularly for chemical molecular data.
To address this challenge, some researchers develop special traversal methods~\cite{singh2020chemoverse, zhang2023graph, zheng2023desirable}.
For example, ChemoVerse~\cite{singh2020chemoverse} introduces a manifold traversal algorithm.
To find a path going through regions of interest, a k-d tree is built based on the Jacobian distances of all points of interest and additional user-specified constraints.
Then $A^*$ algorithm is combined with Yen's algorithm~\cite{yen1971finding} to find the shortest path in the k-d tree.
Subsequently, data interpolation is performed along this path by sampling points at equal interval and decoding the latent vectors into full-fledged molecular structures with the generation model.
\end{itemize}

\subsubsection{Tabular Generation}

\stitle{Data}. GenAI can be used to synthesize surrogate data for subsequent tasks, such as privacy protection.
Specifically, to protect users' data, oftentimes many institutions would not reveal the real data to the public, causing lack of data for domain-specific analysis tasks. 
Instead, some studies aim at generating surrogate data similar to real data which can be freely used to test downstream tasks such as visualization and query~\cite{fan13relational, chen2019faketables, park2018data, qinl2022synthesizing, xu2019modeling, zhang2017privbayes}.
The data are typically relational data.

\begin{itemize}
    \item \emph{Relational data}.
    Relational data is the most basic form of data for visualization which are often stored in tabular format comprising data items in rows and multi-dimensional attributes in columns.
    Surrogate data generation studies mainly focus on tabular relational data.
\end{itemize}

\stitle{Method}. A common method for tabular data generation is GAN.
\begin{itemize}
    \item \emph{Generative Adversarial Network (GAN)}.
    For example, in recent years, some researchers attempt to generate relational data similar to real data with GANs~\cite{fan13relational, park2018data, qinl2022synthesizing, xu2019modeling}.
    The architecture of GANs consists of a generator and a discriminator.
    The adversarial training scheme where the generator progressively learn to generate more realistic data that can deceive the discriminator enables GANs to model the distribution of real data.
    For example, table-GAN~\cite{park2018data} builds upon the basic deep convolutional GAN (DCGAN)~\cite{radford2015unsupervised} framework and tailor the generation to tabular data.
    Specifically, first the tabular records are converted into square matrix to accommodate convolution operation.
    In addition to the original generator and discriminator, table-GAN also incorporates a classifier network which learns the correlation between categorical labels and other attributes from the table.  
    This serves to maintain the consistency of values in the generated records.
    Moreover, besides the original adversarial loss, the authors design a new information loss, which measures the first order and second order statistical difference between the high-dimensional embedding vectors before the sigmoid function in the discriminator.
    However, purely GAN-based methods still leak important features of user data as they are directly trained on real data.
    To address this risk, SERD~\cite{qinl2022synthesizing} seeks to generate similar data while preserving the key privacy information in real data.
    Specifically, SERD manages to satisfy the differential privacy guarantee conditions by using fake entities satisfying the same vectorized similarity constraint of entities in real datasets.
\end{itemize}

\subsubsection{Spatial Generation}\label{sssec:data_spa}

\stitle{Data}. GenAI can be used to infer spatial data such as imagery, volumetric or spatial-temporal data.
\begin{itemize}
    \item 
    \emph{Imagery data}.
    Some studies apply GenAI to image data inference, including emoji images~\cite{liu2019latent}, medical images~\cite{li2021review}, natural images~\cite{huang2018auggan, choi2019self}, etc.
    The inferred data are used for subsequent tasks like visual interpolation~\cite{liu2019latent, liu2018data}, super-resolution~\cite{li2021review}, 3D reconstruction~\cite{wang2019deeporgannet}, object detection~\cite{gou2020vatld} and semantic segmentation~\cite{he2021can}.
    \item 
    \emph{Volumetric data}.
    Volumetric data is a type of data that represents information in three-dimensional space, which is widely used in various fields such as biology, geology, and physics.
    Some studies use GenAI for volumetric data super-resolution to address the problem of low data quality~\cite{zhou2017volume}.
    Others works focus on volumetric data reconstruction through generation model.
    For example, DeepOrganNet~\cite{wang2019deeporgannet} applies GenAI to reconstructing and visualizing high-fidelity 3D organ models based on input of merely single-view medical images.
    
    \item 
    \emph{Spatial-temporal data}.
    Spatial-temporal data such as flow data is a type of data that combines both spatial and temporal components. 
    It involves information that varies not only in space (location) but also over time.
    GenAI can be applied to data extrapolation of spatial-temporal data.
    For example, Wiewel et al.~\cite{wiewel2019latent} leverages GenAI to model the temporal evolution of fluid flow.
    Super-resolution can also work on spatial-temporal data.
    For example, STNet~\cite{han2021stnet} addresses the spatial-temporal super-resolution of volumetric data. 
\end{itemize}

\stitle{Method}. Spatial data inference typically include VAE, GAN and DRL methods. 
\begin{itemize}
    \item 
    \emph{Variational Autoencoder (VAE)}. VAE~\cite{kingma2013auto} is a commonly used generative method that formulates generation as an autoregressive learning framework.
    The basic autoencoder architecture consists of an encoder extracting data features into latent representation vectors and a decoder reconstructing the data from the latent vectors.
    To allow GenAI to capture the variability in data, VAE builds on traditional autoencoder by modeling latent representation as probablistic instead of a fixed vector.
    During generation, the decoder sample from the distribution in the latent space and synthesize new data.
    VAE has been exploited for many data inference tasks in visualization such as data interpolation.
    For example, Latent Space Cartography~\cite{liu2019latent} trains multiple VAE models with different hyperparameters on 24,000 emoji images.
    Users can explore the latent space of these VAEs and define customize semantic axes by selecting samples representing two ends of opposing concepts.
    Subsequently, linear interpolation is performed at constant intervals along the axis to generate intermediate samples showing the transitions of visual features of the emoji images. 
    \item 
    \emph{Generative Adversarial Network (GAN).} GAN~\cite{goodfellow2020generative} models the process of generation as an adversarial learning framework, the basic components of which are the generator and the discriminator.
    The generator $G$ is designed to generate data that resembles real training data.
    The discriminator $D$ is designed to distinguish the data generated by the generator from the real data.
    The training alternates between the generator and discriminator to optimize a min-max problem with the following objective function.
    Different from the original GAN, spatial-temporal adversarial generation requires a spatial-temporal generator and discriminator.
    For example, STNet~\cite{han2021stnet} builds a ConvLSTM structure for discriminator.
    Specifically, Convolution layers are used to extract spatial features.
    Then features of adjacent time steps are fed into ConvLSTM to evaluate temporal coherence.
    Global average pooling is used to produce the final single value score for realness.
    \item 
    \emph{Disentangled Representation Learning (DRL)}. DRL with VAEs or GANs have been applied to visual analytics to identify interpretable dimensions interactively~\cite{gou2020vatld, wang2020scanviz, he2021can, wang2023drava, evirgen2023ganravel}.
    The disentangled dimensions can be subsequently controlled by users to generate meaningful data for augmentation.
    In the general literature of computer vision and computer graphics, disentanglement has been an essential technique for controllable generation~\cite{tran2017disentangled, jeon2021ib, zhou2023clip}.
    A commonly used DRL architecture is $\beta$-VAE~\cite{burgess2018understanding}.
    The objective function of $\beta$-VAE is a modification of the original VAE with an additional $\beta$ parameter.
    Experiments show that better chosen $\beta$ value (typically $<1$) can produce more disentangled latent representation $\mathbf{z}$. 
    For example, VATLD~\cite{gou2020vatld} is a visual analytics system that adapts $\beta$-VAE to extract user-interpretable features like colors, background and rotation from low-level features of traffic light images.
    With such interpretable features encoded in latent space, users can generate additional training examples in an interpretable manner to enhance the traffic light detection model.
    The DRL scheme distills potentially significant semantic dimensions in latent space representation for data summarization and semantic control by users.
    Particularly, two additional losses are introduced to the original $\beta$-VAE, namely the prediction loss and perceptual loss to ensure generation and reconstruction of more realistic traffic light images.
\end{itemize}

\begin{figure*}[t]
\centering
\includegraphics[width=0.995\textwidth]{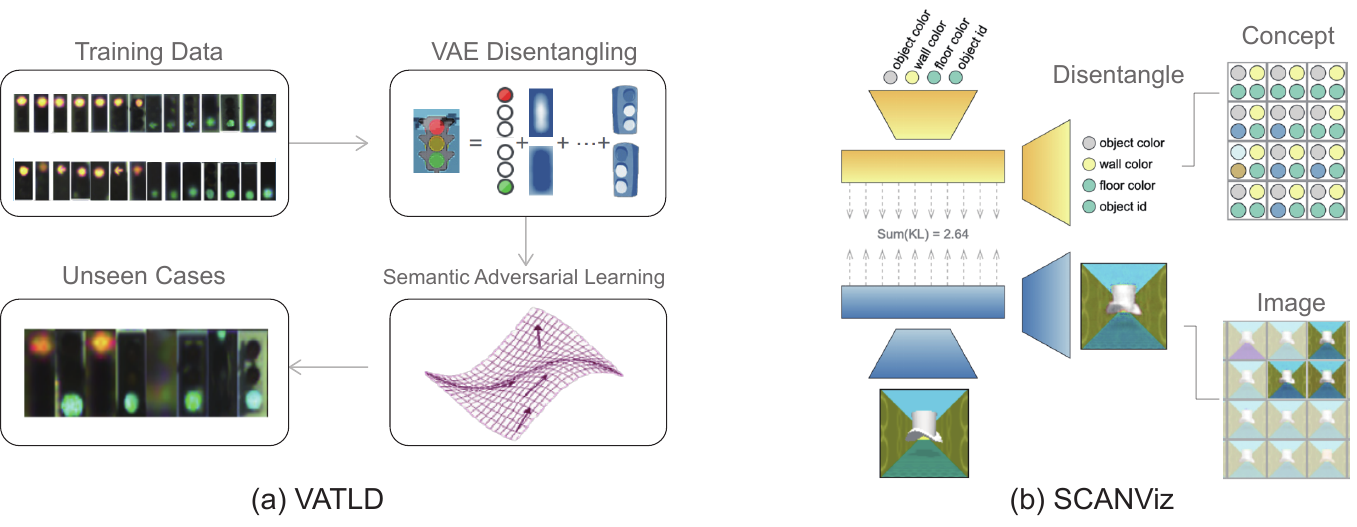}
\vspace{-2mm}
\caption{\revise{Data enhancement with disentangled representation learning, such as VATLD~\cite{gou2020vatld} and SCANViz~\cite{wang2020scanviz}.}
}
\label{fig:vatld}
\end{figure*}

\subsubsection{Discussion}

GenAI methods like GANs are not designed to predict the accurate data values. 
Instead, they focus on generating reasonable data based on the given distribution of the real data.
Such generation should not be overclaimed or misused for tasks that require accurate data features.
A specific example is outlier data, which might pose challenges as GenAI methods predominantly concentrate on learning the overall data distribution.
Particularly, generative models aim to generate data that closely matches the majority of the training data. If the outliers are rare and significantly different from the majority of the data, the generative model may not capture them effectively. Outliers can be overlooked or underrepresented in the generated samples.

Despite being designed to embed data in more disentangled dimensions, the automatic DRL methods cannot guarantee the resulting dimensions are perfectly interpretable and disentangled.
Consequently, the generative models built upon DRL are still largely a black box.
For example, the choice of $\beta$ in $\beta$-VAE still remains largely heuristics.
In addition, even though the dimensions extracted by DRL may be meaningful, it may not encode the visual properties users intend to explore, thus can potentially limit the customizable data exploration on users' part.
Some recent work attempts to incorporate more user interaction to refine DRL through visual interface.
For example, DRAVA~\cite{wang2023drava} not only allows the meaning of DRL dimensions to be verified by users, but also enables user refinement.
To facilitate user refinement of concept dimensions, they propose a light weight concept adaptor network on top of the VAE. 
The concept adaptor is a multi-class classifier to predict the correct grouping of data points along a selected dimensions for semantic clusterings.
However, such interactions are still limited in some aspects.
Users may only be able to verify and refine a small subset of dimensions, leaving many others unaddressed, because of the lack of overview for the relations between data points and different dimensions.

\subsection{Data Embedding}\label{ssec:data_embed}
Data embedding is an emergent technology that leverages GenAI to embed data into visualization images with information steganography.
The data can be recovered losslessly from the visualization images, which are mostly 2D imagery data.

\subsubsection{Spatial Generation}

\stitle{Data}. Spatial generation in data embedding concerns QR code and visualization image data. 
\begin{itemize}

\item
\emph{QR Code}. QR code data is a special type of data used as the visual coding scheme of the chart information such as metadata to be embedded, which can be processed together with the visualization images with neural networks. 
QR code is a reliable coding scheme allowing for error correction but avoiding artifacts in the encoded image~\cite{ye_2023_invVis, zhang2020viscode}.

\item
\emph{Visualization Image}. Although QR code can encode chart information, it has limited data encoding capacity.
For the task of data embedding, visualization images are the essential medium to carry the encoded data and information.
Large number of visualization images are needed to train the model.
Synthetic data can be used for training, but real-world visualization image datasets such as VIS30K~\cite{chen2021vis30k} and MASSVIS~\cite{borkin2013makes} have also been used to increase the generalizability and robustness of the model.
To embed large quantities of underlying data for invertible visualization, data image that represents raw data produced by data-to-image (DTOI) method can also be used~\cite{ye_2023_invVis}.

\end{itemize}

\begin{figure*}[t]
\centering
\includegraphics[width=0.995\textwidth]{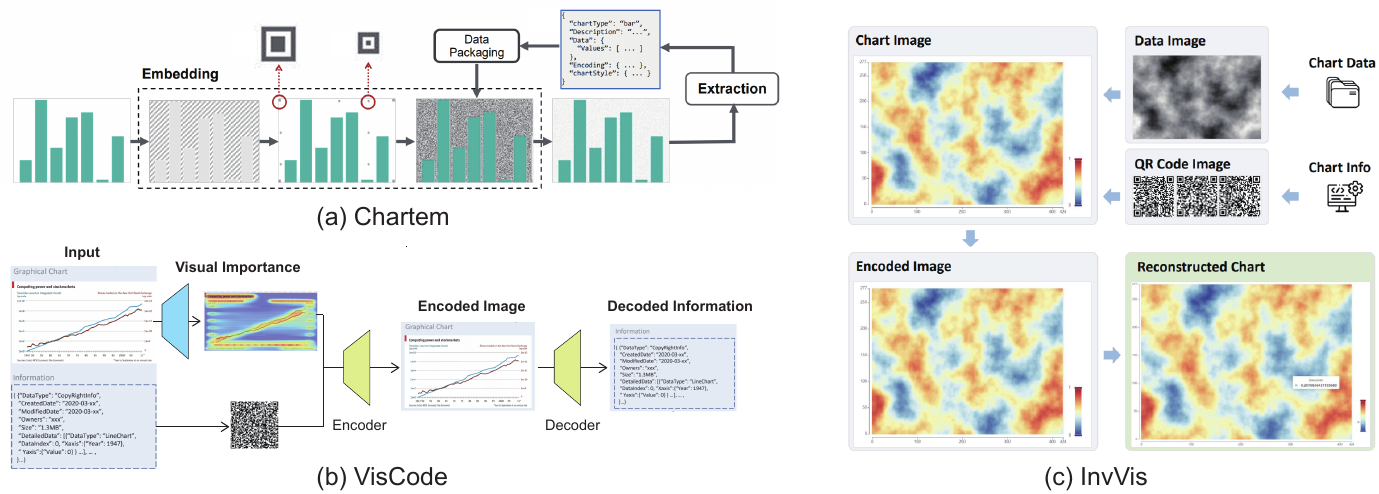}
\vspace{-2mm}
\caption{\revise{The data embedding pipeline of (a) Chartem~\cite{fu2020chartem}, (b) VisCode~\cite{zhang2020viscode} and (c) InvVis~\cite{ye_2023_invVis}.}
}
\label{fig:embed}
\end{figure*}

\stitle{Method}. The data embedding methods include boundary-aware segmentation network (BASNet) and invertible steganography network (ISN) models.

\begin{itemize}
    \item 
\emph{Saliency BASNet}.
In order to assess the visual quality of the coded visualization image to ensure it is perceptually identical to the original image, it is insufficient to measure the pixel-wise mean square error because it neglects the varying importance of pixels across the visualization.
To address this issue, VisCode~\cite{zhang2020viscode} proposes a special visual importance network to predict the visual importance map for the chart image.
Compared to traditional saliency-based method which are applied to natural images but overlook the unique features of visualization images, the visual importance network can learn from eye-movement data of real users on chart images.
Specifically, the model adopts a BASNet~\cite{qin2019basnet} architecture which is originally developed for salient object detection.
The architecture is based on U-Net structure with residual blocks inspired by ResNet.
The loss function combines BCEWIthLogits loss with structural similarity index (SSIM) to balance between segmentation accuracy and structural information.

\item
\emph{Encoder-decoder ISN model}.
To encode and decode the visualization image, QR code image and data image in one unified model, InvVis~\cite{ye_2023_invVis} introduces a concealing and revealing network.
The concealing network and revealing network both consist of two major parts: feature fusion block (FFB) and invertible steganography network (ISN).
FFB is designed to blend the features of data image and QR code image into visualization image while keeping minimal visual distortion.
Specifically, FFB comprises four dense blocks and three common convolutional blocks.
Dense blocks~\cite{huang2017densely} are a special type of convolutional neural network architecture which contain multiple layer with dense connection (each layer is connected to all the preceding layers).
Next, the ISN adds the invertible $1\times1$ convolution to the invertible neural network structure~\cite{dinh2014nice}, which consist of several affine coupling layers.
The authors also proposed using discrete wavelet transform (DWT) between the FFB and ISN to reduce texture-copying artifacts.
\end{itemize}

\subsubsection{Discussion}
Currently, the evaluation of the quality of data restoration is only limited to the data image, using generic metric such as root mean square error (RMSE).
However, such pixel-wise metrics cannot fully reflect the accuracy of data restoration due to the absence of original data in the evaluation.
In addition, the capacity of visualization image for data embedding is not infinite.
Specifically, there can be an apparent trade-off between embedding capacity and image quality.
To maintain image quality above certain threshold, it becomes more challenging to recover large amounts of data.
To address this concern, more evaluation in practical scenario regarding precision requirements for the original data need to be conducted.

\section{Visual Mapping Generation}
\label{sec:vis_map}
For non-expert users, it is difficult to make appropriate visualization from data on their own for data analysis.
GenAI plays an essential role in visual mapping synthesis for automatic visualization generation.

\subsection{Visual Primitives Generation}
The fundamental task of visual mapping generation is generating charts with the basic visual marks or visual primitives. 

\subsubsection{Sequence Generation}

\stitle{Data}. Sequence generation is often applied to visual mapping in different visualization grammars including concrete programming language and abstract code.
\begin{itemize}
\item
\emph{Visualization grammar}.
Generative AI can be applied to generate visualization code in different grammars based on input data.
For example, Vega-Lite code is a commonly used declarative visualization language~\cite{dibia2019data2vis, narechania2020nl4dv}.
Data2Vis~\cite{dibia2019data2vis} treats generation of visual encoding as a sequence-to-sequence generation task which translate strings describing columns of data tables into Vega-Lite code sequences.
Other online repositories such as Plotly also provides codes in other programming languages such as python.
In addition, some studies generate abstract code instead of specific programming code.
For example, Table2Charts~\cite{zhou2021table2charts}, the authors define a more abstract chart template language including the essential visual elements and a set of grammar that summarizes the possible actions in the process of chart creation.

\end{itemize}

\stitle{Method}. Sequence generation used in visual mapping typically includes RNN, Deep Q Network, NL2VIS Transformer and the latest LLMs.

\begin{itemize}

\item
\emph{RNN-based Code Sequence Generation}.
Some studies formulate the problem of visualization code generation as sequence-to-sequence generation.
For example, Data2Vis~\cite{dibia2019data2vis} translates strings describing columns of data tables into Vega-Lite code sequences. 
For this task, the authors take inspiration from machine translation and adopt a encode-decoder architecture based on recurrent neural network model.
Specifically, for the decoder, they construct a two-layer bidirectional RNN;
for the decoder, another two-layer RNN is used to predict the next token in the code sequence.
Both the encoder and decoder leverage the Long Short Term Memory (LSTM) structure to enhance the model's ability to deal with longer sequence.

\item
\emph{Deep Q network for encoding action prediction}.
Some studies consider the task of visualization generation as the generation of abstract action tokens deciding key features of the charts, including data queries which select particular fields in data table and design choices which specify visual encoding operation.
In this light, visualization generation can be formulated as an action prediction task which can be solved by Deep Q Network (DQN).
For example, Table2Charts~\cite{zhou2021table2charts} develops a simple chart template language describing some essential actions for generation of six types of charts from data tables.
According to this template, the authors construct a DQN for action prediction with a customized CopyNet architecture~\cite{gu2016incorporating}.
This network takes all the data fields and prefix action sequence as input and generate the next action token with a Gated Recurrent Unit (GRU) based RNN structure.
In additon, to address the exposure bias problem with the previous teacher forcing training scheme which only learns the ground truth user generated results, Table2Charts adopts the search sampling approach of reinforcement learning to close the gap between training and inference.

\item
\emph{Natural language to visualization models}.
The Natural language to Visualization (NL2VIS)~\cite{ADVISor, ncnet, nvbench, song2022rgvisnet} task can be formulated as follows:
Given a natural language query ($NL$) over a dataset or relational database ($D$), the goal is to generate a visualization query (\eg~Vega-Lite) that is equivalent in meaning, valid for the specified $D$, and, when executed, will return a rendered visualization ($VIS$) result that aligns with the user’s intent.
For example, ADVISor~\cite{ADVISor} trained two separate neural networks to provide the NL2VIS functionality.
Broadly, ADVISor's pipeline can be divided into two steps: (1) the NL2SQL step, and (2) the rule-based visualization generation step.
ADVISor takes as input a NL question and data attributes associated with the datasets. Next, it first utilizes a BERT-based neural network to generate a vector representation ($q$) of the NL question  and a corresponding header vector.
Subsequently, the system utilizes an Aggregation network to deduce aggregate operators and a Data network to determine attribute selection and filtering conditions. Upon completing these steps, ADVISor first queries the dataset. It then maps the query results to a visualization based on a rule-based visualization algorithm
The neural network modules within ADVISor are specifically trained to extract fragments of SQL queries from the given NL query, which indicates  that ADVISor is not an end-to-end NL2VIS solution.
On the contrary, ncNet~\cite{ncnet} is an end-to-end NL2VIS method based on the Transformer~\cite{transformer} architecture. ncNet is trained on the first large-scale cross domain NL2VIS dataset nvBench~\cite{nvbench}.
ncNet utilizes a Transformer-based encoder-decoder framework, with both the encoder and decoder comprising self-attention blocks. This system accepts a NL query, a dataset, and an optional chart template as inputs. ncNet processes these inputs into embeddings and finally generates a flattened visualization query through an auto-regressive mechanism. Additionally, ncNet incorporates a visualization-aware decoding strategy, which allows for the generation of the final visualization query, with visualization-specific knowledge.

\item
\emph{Large language model for visualization code generation}.
Recently, some researchers realize the limitation of previous GenAI methods which only focus on a particular type of visualization code like Vega-Lite.
To improve the flexibility of visualization code generation, some studies propose using large language model for more robust generation~\cite{dibia2023lida, wang2023llm4vis, li2024prompt4vis, tian2024chartgpt, li2024visualization}.
For example, LIDA~\cite{dibia2023lida} presents a pipeline called VISGENERATOR for AI generation of grammar-agnostic visualizations connecting data tables to generated visualizations with multiple steps.
The VISGENERATOR consists of three sub-modules: code scaffold constructor, code generator and code executor.
The code scaffold constructor generates code that imports language-specific dependencies like Matplotlib and constructs the empty function stub.
Then, in code generator, taking as input the dataset summary and visualization goal, LLM is used in the fill-in-the-middle mode~\cite{bavarian2022efficient} to generate concrete visualization code of the given programming language.
Finally, in the code executor, some filtering mechanisms such as self-consistency~\cite{wang2022self} and correctness probabilities~\cite{kadavath2022language} are incorporated to reduce errors.
In the third step of LIDA, taking as input the dataset summary and visualization goal, LLM is used in the fill-in-the-middle mode~\cite{bavarian2022efficient} to generate concrete visualization code of different programming languages.
Another recent study, LLM4Vis~\cite{wang2023llm4vis} proposes leveraging the in-context learning ability of large language models to perform few-shot and zero-shot generation for the same design choice task as in VizML~\cite{hu2019vizml}.
The key contribution of this method compared to previous supervised learning is that it reduces the need for large corpus of data-visualization pair training data and provides more explainable generation.
The generation algorithm is retrieval-augmented with demonstration examples.
First, data feature description is generated  to enable GPT to take tabular dataset as input.
Specifically, similar to VizML, with feature engineering as many as 120 single-column features and 80 cross-column features are extracted to represent the input dataset.
These features are then serialized by TabLLM method~\cite{hegselmann2023tabllm}, which utilizes a prompt 
to instruct ChatGPT to generate text description that elaborate on the feature values for each attribute.
Next, the demonstration examples are selected from the training corpus by similarity retrieval based on feature description, which can fit into token limit of LLM without considering irrelevant samples.
Subsequently, an iterative explanation generation bootstrapping module prompts LLM to not only predict the correct visual design choices but also generate explanation.
Finally, all the relevant demonstration examples along with the explanation are fed into LLM to optimize in-context learning for generation of appropriate design choices for the input data.
Recently some researchers also leverage LLM to refine the colormap~\cite{shi2023nl2color}.

\end{itemize}

\subsubsection{Tabular Generation}

\begin{figure*}[t]
\centering
\includegraphics[width=0.995\textwidth]{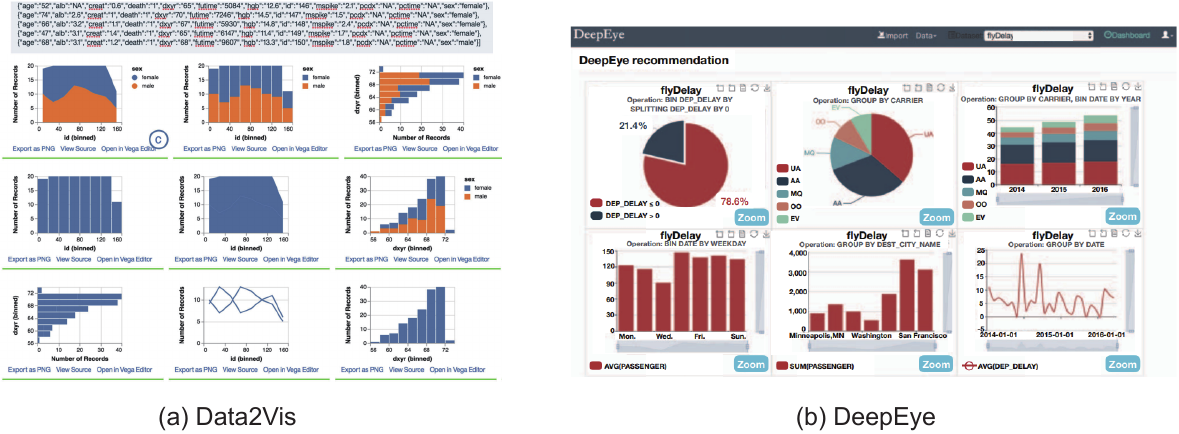}
\vspace{-2mm}
\caption{\revise{Examples of visual primitive generation. (a) Data2Vis~\cite{dibia2019data2vis} adopts RNN-based sequence generation method. (b) DeepEye~\cite{luo2018deepeye} combine design parameter enumeration with AI scoring.} }
\label{fig:vismap}
\end{figure*}

\stitle{Data}. The visual mapping can also be simplified as predicting some tabular attributes such as design choices.

\begin{itemize}
\item
\emph{Design choices}. 
Instead of directly generating the visualization images, some studies focus on generating the most important design choices for the specifications of visualization, which is represented as tabular data with each attribute denoting one design dimension.

\end{itemize}

\stitle{Method}. The tabular generation methods of design choices include fully connected neural network for direct prediction and design parameter enumeration with AI scoring.

\begin{itemize}

\item
\emph{Fully connected neural network} can be combined with feature engineering of data, casting the generation problem into a prediction task.
For example, VizML~\cite{hu2019vizml} builds a feed-forward neural network based on 841 dataset-level features extracted from input data table.
The model predicts five design parameters of the appropriate visualizations with multi-head output layer, on both the encoding level and visualization level.
For example, to generate the visualization type, one of the prediction head outputs a 6-class prediction scores for chart types including \emph{Scatter, Line, Bar, Box, Histogram, Pie}.

\item
\emph{Design parameters enumeration with AI scoring}.
Some studies approach the generation problem from a different angle, using AI as the judge of the candidate generation results.
For example, DeepEye~\cite{luo2018deepeye, luo2018deep, luo2020steerable, DBLP:journals/bigdatama/QinLTL18, DBLP:conf/edbt/QinL0018} combines rule-based generation with data-driven machine learning to classify and rank meaningful visualizations.
Specifically, based on a collected corpus of real world visualization cases, a classification model and a ranking model are trained.
When users input a new dataset to visualize, the system first generates candidate visualizations by enumerating valid combinations of transformations and visual encoding in a pre-deifined search space.
The classification model then determines whether a visualization candidate is meaningful.
Subsequently, the ranking model sort the remaining meaningful visualization and recommend to users.
In case the machine learning models do not yield satisfactory results, DeepEye also supports incorporation of expert-designed domain rules.
The recommendations of the machine learning model and the rule-based method can also be combined with a linear model.
Similarly, Text-to-viz~\cite{cui2019text} adopt a hybrid method combining template-based enumeration with AI-powered comprehension of user inputs and relevance ranking.

\end{itemize}

\subsubsection{Spatial Generation}

\stitle{Data}. Spatial generation in visual mapping mainly concerns 2D sketch, density map and volumetric data.

\begin{itemize}
\item
\emph{Sketch}.
Sketch is a special type of 2D image data with simple information of drawing trajectory.
It has attracted much research interest in the broader field of generative AI~\cite{voynov2023sketch} because it allows designers to follow their familiar workflow of prototype design and allows them to have spatial control over the generated results.
Recently, some studies explore using sketch as a medium to facilitate fast prototyping of visualizations with sketch-to-vis generative AI~\cite{ma2020ladv, teng2021sketch2vis}.

\item
\emph{Density map}. 
Density maps are a special type of visualization that can vary dynamically depending on the time.
Traditionally, the spatial temporal data collected for density map visualization is discrete and static.
For smoother transition of density maps at different discrete observation times, some researchers propose to utilize generative AI~\cite{chen2019generativemap}.

\item
\emph{Volumetric data}. Apart from the enhancement of volumetric data as introduced in Section~\ref{sssec:data_spa}, some studies also leverage GenAI methods for the rendering of such data~\cite{berger2018generative, hong2019dnn}.

\end{itemize}

\begin{figure*}[t]
\centering
\includegraphics[width=0.995\textwidth]{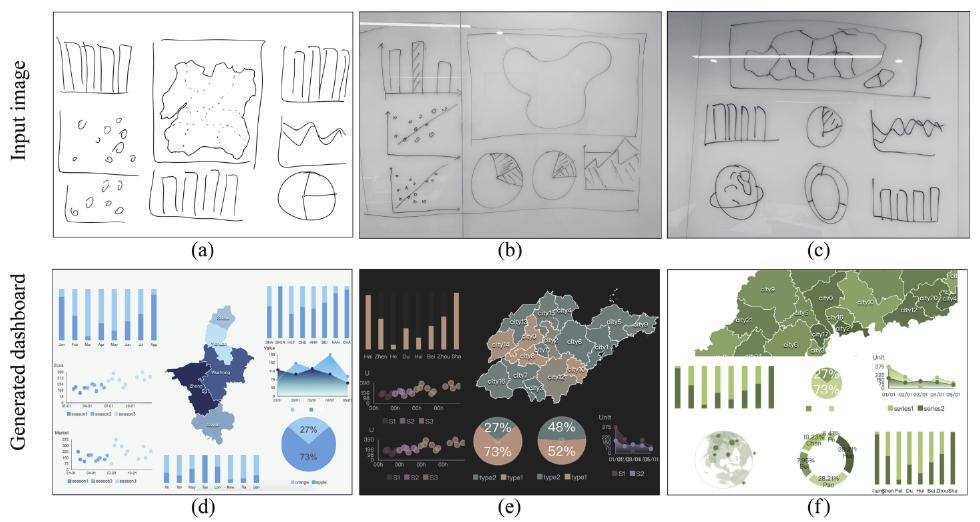}
\vspace{-2mm}
\caption{\revise{Examples of LADV~\cite{ma2020ladv} sketch to dashboard generation.}}
\label{fig:sketch}
\end{figure*}

\stitle{Method}. Spatial generation methods for visual mapping mainly includes Faster R-CNN with validation model, GAN-based density map generation and GAN-based volume rendering.

\begin{itemize}

\item
\emph{Faster R-CNN for sketch recognition with validation model}.
For the task of sketch to dashboard generation, latest AI-driven approach combines AI-powered chart recognition with rendering algorithms such as color palette recommendation and layout optimization.
To detect the charts and the basic visual encoding features, LADV~\cite{ma2020ladv} first applies a Faster R-CNN network~\cite{ren2015faster} which exploits region proposal to achieve efficient object detection.
Specifically, Faster R-CNN further accelerates Fast R-CNN by computing the region proposal with a deep CNN-based region proposal network which can share weights with the subsequent object detection network.
In addition, to adapt Faster R-CNN to chart recognition, LADV~\cite{ma2020ladv} further incorporates a validation model to filter chart candidates, which trains a logistic regression for each chart type to capture the location and size.

\item
\emph{GAN-based density generative model}.
To generate dynamic density maps, density generative model is developed.
For example, GenerativeMap~\cite{chen2019generativemap} adopts a GAN-based generative model.
Specifically, the authors first generate synthetic training data using Perlin noise.
Then, they adapt Bidirectional Generative Adversarial Network (BiGAN)~\cite{donahue2016adversarial} to the density generation with enlarged convolution kernels and blocks inspired by ResNet~\cite{he2016deep} to enable processing of larger fields.
Poisson blending is also used to make the density change more natural.

\item
\emph{GAN-based Volume rendering}.
Some researchers adopt GenAI for the rendering of 3D volumetric data~\cite{berger2018generative, hong2019dnn, wu2022diver, gan2023v4d}.
For example, Berger et al.~\cite{berger2018generative} proposed a framework that combines two GANs for the task, namely opacity GAN and opacity-to-color translation GAN. 
Such approach breaks down the more difficult task of rendering volumes compared to the image generation task of the original GAN. 
Specifically, the opacity GAN learns to generate an opacity image given the input of viewpoint and opacity transfer function, which captures the shape, silhouette, and opacity.
The second GAN translates the combined inputs of the viewpoint, the opacity transfer function’s representation in the latent space, color transfer function, as well as the opacity image generated by the first GAN into the final rendered image.

\end{itemize}

\subsubsection{Graph Generation}
\begin{figure*}[t]
\centering
\includegraphics[width=0.995\textwidth]{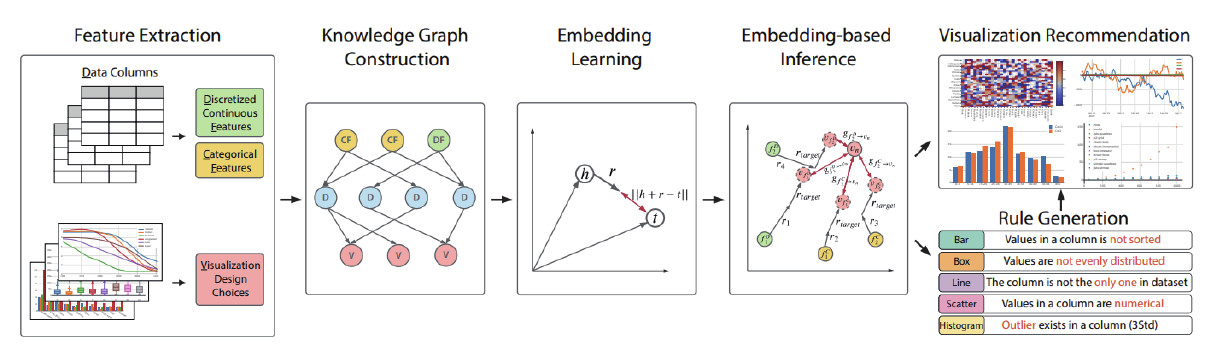}
\vspace{-2mm}
\caption{\revise{Knowledge-graph-empowered visual mapping generation in KG4VIS~\cite{li2021kg4vis}.}}
\label{fig:kg}
\end{figure*}

\stitle{Data}. The graph generation in visual mapping mainly involves knowledge graphs.
\begin{itemize}
\item 
\emph{Knowledge graph}.
Knowledge graph is a graph structure representation of knowledge that captures the relationships between entities in a particular domain, which is composed of nodes representing entities and edges representing relationships between these entities.
Some studies also leverage knowledge graphs to support more explainable generation of visualization~\cite{li2021kg4vis}.

\end{itemize}

\stitle{Method}. The method for knowledge graph enhanced visual mapping generation is knowledge graph embedding.

\begin{itemize}
\item
\emph{Knowledge-graph embedding}.
KG4VIS~\cite{li2021kg4vis} applies knowledge graph to the design choice generation task of VizML~\cite{hu2019vizml}.
The knowledge graph is constructed with entities representing data features, data columns and visualization design choices, and the relations between them.
Then, KG4Vis employs knowledge graph embedding method TransE~\cite{sun2018rotate} to learn neural embedding representations of all the entities and relations.
When generating design choices for new input data, the model encode the input data in the embedding space and evaluate the scores of different candidate visual encoding rules with arithmetic operation adding relation vectors to entity vectors and measuring the distance from design choice vectors.
\end{itemize}

\subsubsection{Discussion}
\textbf{Limitations of the rules and the training data}.
The training data can be a significant limiting factor for visual primitive generation.
For example, Data2Vis~\cite{dibia2019data2vis} only trains the model on four major chart types, which significantly limits the scope and diversity of the generation.
In addition, some hybrid methods combining rule-based components with GenAI may be limited by the rules that are not comprehensive enough.
For example, the chart template defined in Table2Charts only~\cite{zhou2021table2charts} include the most basic visual mark and direct data field reference without even considering some basic operation like aggregation.

\textbf{Generation of visualization images vs. natural images}. Image generation has been a heavily researched topic in the general field of AI and computer vision.
However, the generation of visualization images rarely adopts a fully end-to-end method without rule-based constraints.
This is partly because of the difference between visualization images and natural images.
Especially, compared to the irregular and complex visual features in natural images, the dominance of regular shapes rigid structures in visualization images makes it difficult for GenAI to accurately maintain.
The reason is AI models are inherently stochastic and treats the image features as a whole with little knowledge of the structural constraints.
To address this issue, future work may take inspiration from some recent computer vision studies that seek to incorporate structural information in image processing~\cite{kai2022visiongnn}.

\textbf{3D diffusion}.
Recently, inspired by the success of text-to-image generation diffusion model, some researchers seek to develop diffusion-based text-to-3D~\cite{poole2022dreamfusion, sella2023vox} models to allow more intuitive interactive control of the generation, including natural language guided generation and editing.
Such technology can potentially benefit generation of visualization, especially in providing multi-modal control.

\textbf{NL2VIS challenges}. 
Unquestionably, LLMs offer a complementary dimension to the NL2VIS system. However, the integration of LLMs into NL2VIS through prompt engineering presents certain limitations.
First, relying solely on a simple prompting-based method may not effectively enable LLMs to fully comprehend the intricacies of the NL2VIS task. This approach could limit the model's ability to accurately interpret and respond to more complex visualization queries, potentially overlooking nuanced aspects of data visualization requirements.
Second, current LLMs-based NL2VIS solutions often do not incorporate specific domain knowledge from the fields of visualization and data analysis into the LLMs. This absence of domain-specific integration can result in suboptimal performance, as the models may not leverage the rich contextual and technical knowledge necessary for producing highly accurate and relevant visualizations.
Furthermore, current LLMs-based NL2VIS solutions struggle to guarantee the semantic correctness of generated visualizations, which is crucial for accurate data representation and interpretation. Additionally, these systems often face challenges in interactively fine-tuning results based on user feedback, a key aspect for achieving user-centered visualization design.

Given these challenges, future research directions may include:
1) Developing methods to integrate specific domain knowledge related to visualization and data analysis into LLMs.  This could involve training models on specialized datasets or incorporating expert systems that guide the LLMs in understanding domain-specific visualization tasks, knowledge, and requirements.
2) Ensuring the semantic correctness of visualizations generated by LLMs, which can explore validation strategies that automatically check and confirm the accuracy and relevance of visualizations with respect to the underlying data.
3) Enhancing the interactivity of LLMs-based NL2VIS systems by incorporating more robust and flexible user feedback loops. For example, it can explore how to incorporate other modalities (\eg clicks) user feedback.
4) Investigating hybrid models that combine the strengths of LLMs with traditional data visualization techniques. Such hybrid systems could leverage the natural language understanding capabilities of LLMs while ensuring adherence to best practices in data visualization.

\section{Stylization}
\label{sec:style}

\subsection{Style Transfer}

Style broadly refers to the visual or aesthetic characteristics of an image, which oftentimes involves some global or overall features that can affect the viewers' general appreciation.
More concretely, it can involve many aspects such as color, texture and layout.
Some style transfer studies focus on the overall style while others focus on particular aspects like color.

Creating visualization in design practice relies on existing examples from the internet, offering inspirational visual materials toward style. This has spawned extensive research on transferring style attributes from these existing materials to facilitate the development of intended visualization designs.
To transfer overall visual style, some research~\cite{harper2017converting, huang2021visual, shi2022supporting, ma2020ladv} summarize style as a template and migrate these graphical attributes disentangled from the content to restyle new data source.

\subsubsection{Spatial Generation}

\stitle{Data}.
Researchers frequently employ visualizations in image format to train GenAI models, which are designed to extract specific attributes for transfer tasks.

\begin{itemize}
    \item \textit{Imagery data.}
     End-to-end GenAI models consume a substantial amount of visual images, primarily sourced from the internet or synthesized using tools like D3~\cite{bostock2011d3}. Examples include MassVis~\cite{borkin2013makes}, InfoVIF~\cite{lu2020exploring}, and Visually29K~\cite{bylinskii2017learning}.

    \item \textit{Task-oriented Labeling.}
    General image datasets often lack paired attributes, necessitating an extraction stage for labeling task-oriented attributes. For instance, the color transfer task requires extracting the color map for subsequent training. Achieving the overall transfer necessitates considering multiple attributes to describe a comprehensive visual representation.
    
\end{itemize}

\stitle{Method}. The methods include color transfer and hybrid attribute transfer.
\begin{itemize}
    \item \textit{Color transfer.} 
As one of the most important visual channels in data visualization, generations of scholars have investigated the issue of color transfer~\cite{yuan2021deep, yuan2021infocolorizer}.
To extract the color at multiple scales in the Lab histograms, Yuan et al.~\cite{yuan2021deep} employed a neural network featuring an atrous spatial pyramid structure, predicting the colormap of visualization and supporting discrete and continuous formats.
Similarly, Huang et al.~\cite{huang2021visual} approached the problem of color extraction in the foreground of visualization with Faster-RCNN.
With the reference example as the natural image, some research~\cite{shi2022colorcook, liu2022image} generates a harmonious palette for a visualization based on color detection and extraction from images.
For instance, Liu et al.~\cite{liu2022image} distinguished the salient subject in the image to extract the color with high visual importance that aligns with human perception.

\item \textit{Hybrid attribute transfer with Siamese Network}.
Transferring multiple attributes from an example to the current design involves recognizing the content of the example and adapting it to the current design~\cite{shi2022supporting}. Lu et al.~\cite{lu2020exploring} curated a comprehensive infographic dataset and proposed a model based on YOLO to identify various visual elements, including text, icons, indices, and arrows.
To maintain style consistency between the example and the current design, Vistylist~\cite{shi2022supporting} employs a Siamese Neural Network~\cite{lagunas2019learning}. This network embeds visual elements into a 256-dimensional vector and compares the Euclidean distance between pairs. However, assessing the priority of different visual elements in a visualization poses a challenge. 
Huang et al.~\cite{huang2021visual} proposed a restyling approach with an attention mechanism to weigh different visual properties for input visualizations.
Accurate recognition of the example's content and the ability to reproduce it enable hybrid attribute transfer. This not only maintains style consistency but also generates a design tailored to the provided content. For example, when the template timeline has limited space, Chen et al. generated an extended timeline with a similar visual style to the template~\cite{chen2019towards}.
\end{itemize}

\subsubsection{Graph Generation}
\stitle{Data}.
When considering the structure and imagery format of graphs, the data fed into GenAI models falls into two primary categories.

\begin{itemize}
    \item \textit{Graph Structure Data.} 
    This includes nodes and edges in a graph, with the node feature vector and adjacency vector utilized to describe the graph structure that can be recognized by GenAI models. Various embedding techniques, such as node2vec~\cite{grover2016node2vec}, have been proposed to encode node information. 

    \item \textit{Imagery data.} Given that many graphs are in pixel format, GenAI models also process such data to extract low-level features for training.
    
\end{itemize}

\stitle{Method}. The method mainly includes graph layout transfer.
\begin{itemize}
\item \textit{Graph layout transfer.} Some researchers use generative AI for graph layout transfer~\cite{kwon2019deep, wang2019deepdrawing, song2022vividgraph} which seeks to learn the style of graph layout from examples.
For example, to help users intuitively produce diverse graph layouts from a given set of examples without mannually tweaking layout parameters, Kwon and Ma~\cite{kwon2019deep} designed a GenAI method based on encoder-decoder architecture combined with a 2D latent space.
The model generally follows a VAE framework, taking as input graph layout features represented as relative pairwise distances and adjacency matrix.
Then, GNN is used to compute graph-level representation of layout in both the encoder and decoder.
In addition, the latent representation of layout $z_L$ is combined with node-level features through a fusion layer.
The latent space is also visualized in a 2D map to facilitate exploration.
\end{itemize}

\subsubsection{Discussion}
\textbf{Domain knowledge}. Different visualizations have specific requirements and constraints for style transfer, often involving the integration of domain knowledge.
For the categorical data, Zheng et al.~\cite{zheng2022image} introduced a method to sample dominant colors from the image to preserve color discriminability, effectively enhancing and aiding in the interpretation of the patterns present.
As for scientific visualization like terrain maps, domain-specific elements including continuity of elevation and hypsometric tints in aerial perspective are injected into the transfer process to convey the necessary information in a scientifically accurate manner~\cite{wu2023adaptive}.

\textbf{Reference image}. Furthermore, it is worth mentioning that the reference images used for style transfer are not limited to visualization examples alone. Significant works~\cite{poco2017extracting, yuan2021deep, huang2021visual} have explored using natural images as reference sources for style transfer, taping into the inherent visual appeal and cognitive stimulation provided by natural images.
Recent research has shown that natural images can also serve as an adorable source to stimulate human intelligence~\cite{liu2022image, wu2023adaptive}.

\subsection{Chart Embellishment}
Visually embellished visualization showcases its memorability and expressiveness~\cite{borgo2012empirical,harrison2015infographic,haroz2015isotype}. The creation of visually appealing and informative graphical enhancements necessitates design expertise. Fortunately, the advent of generative AI offers a strong framework to streamline the design process, particularly for pictorial visualization and glyph generation.

\subsubsection{Spatial Generation}

\stitle{Data}.
The data that is concerned in GenAI for pictorial visualization is images with semantic correlation with the charts.

\begin{itemize}

\item \textit{Image as pictorial embellishment}. 
In visualization, non-visualization images such as natural images or artistic images can be used as pictorial embellishments to enhance the visual appeal of the data being presented. Images can be added to charts, graphs, and other visualizations to provide additional context and meaning to the data~\cite{coelho2020infomages, zhang2020dataquilt}. For example, an image of a product can be added to a sales chart to help viewers understand which product is being represented by each data point. Similarly, an image of a city skyline can be added to a map to help viewers identify the location being represented.

\end{itemize}

\stitle{Method}. The GenAI method for pictorial visualization mainly include Stable diffusion based techniques.

\begin{itemize}

\item \textit{Stable diffusion}.
Pictorial visualization plays a crucial role in seamlessly integrating semantic context into a chart. Instead of relying on pre-existing graphical elements sourced online and adjusting them to fit the desired visualization, generative AI takes an end-to-end approach by incorporating users' prompts.
Recent advancements~\cite{dibia2023lida, xiao2023let, wu2023viz2viz} in this field have led to the automation of the creation process through the transformation of chart components into semantic-related objects. For example, viz2viz~\cite{wu2023viz2viz} develops specific pipelines for generating various types of visualizations.
The way they leverage textual prompts as input surpasses the previous language-oriented creation tools limited by a predetermined set of entities~\cite{cui2019text, shi2020calliope}, providing a robust semantic recognition to arbitrary user input.
Furthermore, Xiao et al.~\cite{xiao2023let} proposed a unified approach that classifies visual representations into foreground and background. This approach provides users with an interface to select the intended data mask and incorporates an evaluation module to assess the visual distortion in the generated visualization.
\end{itemize}

\subsubsection{Discussion}
\textbf{Flexibility and controllability}. When comparing generative AI with the traditional methods of designing visual elements from scratch or retrieving relevant resources, generative AI offers several distinct advantages. It provides inspiration and eliminates the tedious and time-consuming process of adjusting elements to fit the data. Moreover, it empowers users by allowing them to personalize the style of the generated results with a simple utterance, saving significant time that would otherwise be expended on searching for appropriate resources across the vast expanse of the internet.
Recent advancements in generative models, particularly in the field of text-to-image models, have achieved remarkable breakthroughs in enhancing control over the generated output, including layout~\cite{cheng2023layoutdiffuse}, text content~\cite{chen2023textdiffuser}, vector graphics~\cite{zhang2023text}, etc. 
However, apart from general control, it is imperative to prioritize data integrity throughout the generation process, as visualizations serve to convey data patterns faithfully.

\subsection{Textual Annotation}
Textual annotation plays a pivotal role in the realm of data visualization, enhancing human interpretation, interaction, and comprehension.
Text annotations incorporated in the visualization guide users’ interactions with the artifact~\cite{narrative_2010}, explain what the data means~\cite{graphics_2012}, and prioritize certain interpretations of the data~\cite{hullman_2011}.
In this way, annotations act as cognitive aids, enhancing the overall user experience.

\subsubsection{Sequence Generation}
\stitle{Data.} The data in this task mainly includes text-integrated data and and contextually interpreted data. 
\begin{itemize}
    \item \emph{Text-integrated data.}
    Text-integrated data refers to datasets that inherently come with accompanying textual content, such as articles, reports, or storytelling contexts. In these instances, the data is intertwined with text, forming a cohesive narrative or explanatory structure.
    Textual annotations in such datasets help make complex data more accessible and understandable to the audience, guide their attention to trends that align with textual content, and provide a richer and more nuanced understanding of the data and its relevance to the text.
    \item \emph{Contextually interpreted data.}
    For the datasets that lack accompanying textual narratives or descriptions, the challenge lies in analyzing the data to extract relevant insights and generate meaningful textual annotations that align with the visual elements.
    Effective textual annotations act as a bridge between complex datasets and audiences, providing a layer of interpretation that the raw data lacks on its own.
\end{itemize}

\stitle{Method.}
To further automate the annotation generation process, researchers have developed innovative approaches, mainly through sequence generation. 
\begin{itemize}
    \item 
    \emph{Deep learning detection with template-based generation.} 
    Lai et al.~\cite{automatic_2020} employ a Mask R-CNN model to identify and extract visual elements in the target visualizations, along with their visual properties. The descriptive sentence is displayed beside the described focal areas as annotations.
    AutoCaption~\cite{liu2020autocaption}, a deep learning-powered scheme, generates captions for information charts by learning the noteworthy features aligned with human perception and leveraging one-dimensional residual neural networks to analyze relationships between visualization elements. 
    These advances in automated annotation generation hold promise for applications in education and data overviews.
    Researchers in other areas such as NLP also study the problem of chart summarization~\cite{kantharaj2022chart}. 
    \item \emph{Large Language Models (LLMs).} Recent developments in large language models (LLMs) have opened up new possibilities in generating engaging captions for generic data visualizations.
    Liew and Mueller~\cite{liew2022using} apply LLMs to produce descriptive captions for generic data visualizations like a scatterplot.
    Ko et al.~\cite{ko2023natural} introduce a large language model (LLM) framework to generate rich and diverse natural language (NL) datasets using only Vega-Lite specifications as input.
    This underscores the growing role of prompt engineering techniques in shaping the future of annotation techniques and prompts a reevaluation of research directions.
\end{itemize}

\begin{figure}
    \centering
    \includegraphics[width=0.8\textwidth]{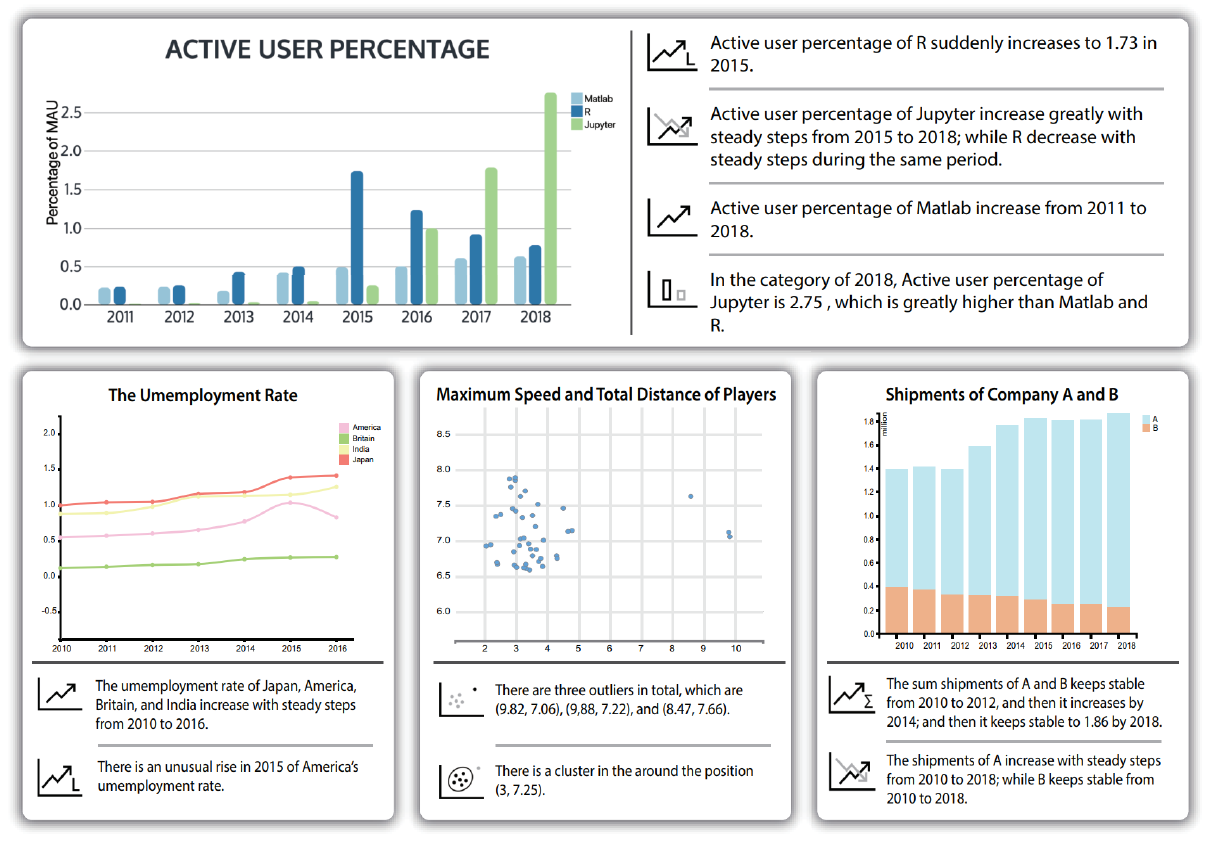}
    \caption{\revise{Annotated example cases of AutoCaption~\cite{liu2020autocaption}. The chart type includes the bar chart, the scatterplot, and the line chart.}}
    \label{fig:autocaption}
\end{figure}

\subsubsection{Discussion}
\textbf{Challenges for textual annotation}. Researchers have consistently underscored the significance of annotations in visualization design, both at the visual memory level~\cite{borkin2013makes} and the cognitive level~\cite{cognitive_2021}.
These studies reaffirm that annotations play an indispensable role in enhancing comprehension and retention of visual information~\cite{chen2023does}.
In the realm of annotation for visualization, it is imperative to address three significant aspects for future research and development: mitigating occlusion problems, harnessing advanced techniques for automation, and enriching visual design.
Firstly, a persistent challenge is the occlusion problem, wherein the annotations block the charts.
Despite considerable efforts to improve the layout of annotated charts, this issue continues to hamper the effectiveness of annotations.
Therefore, more suitable design space should be surveyed and innovative strategies for annotation placement should be considered.
Secondly, recent advancements in deep learning, exemplified by large language models and diffusion models, offer remarkable potential for improving the efficiency of the automatic annotation process.
These models can consider contextual information to produce annotations that are contextually relevant and strategically placed, alleviating the burden on users.
Moreover, it is essential to enhance the design of annotations by incorporating rich visual cues, which aid in highlighting patterns and facilitating deeper understanding.
Researchers should ensure the visually engaging annotations complement the overall visual composition and improve user engagement.

\subsection{Timeline \& Story}

Both timelines and storylines use lines to describe a sequence of events.
Specifically, in a storyline visualization, each role is represented as a line.

\subsubsection{Spatial Generation}
\stitle{Data}. GenAI for story generation commonly concerns about optimizing the spatial layout of different story components.
\begin{itemize}
    \item \emph{Storyline Layout Data}. To training GenAI models to optimize the storyline layouts, it is necessary to generate a large number of high-quality storyline images. In PlotThread~\cite{tang2020plotthread}, Tang \etal construct a dataset of automatically-generated storyline layout pairs, consisting of a original layout and a optimized layout simulated by the optimization model with randomly-selected constraints. 
    
\end{itemize}

\stitle{Method}. The method includes
\begin{itemize}
    \item \emph{Reinforcement Learning for image-based storyline}. To boost a collaborative design of storylines between AI and designers, Tang \etal \cite{tang2020plotthread} further proposed a reinforcement learning framework and introduced an authoring tool, PlotThread, that integrates an AI agent for efficient exploration and flexible customization. The goal of this is to imitate and improve users' intermediate results when optimizing storyling lines. Therefore, it is necessary to understand the states of different layouts, decompose a storyline into a sequence of interactive actions, and provide subsequent actions for layout optimization. They also define the reward as the similarity between the user layout and generated intermediate layout to improve the agent's prediction ability.
\end{itemize}

\subsubsection{Graph Generation}
\stitle{Data}. Researchers seek to understand visualization content automatically from the image input, which can be mainly categorized into raster image and vector image. 
\begin{itemize}
    \item \emph{SVG}. A recent attempt is to apply GenAI technologies to vector charts due to the need of motioning images~\cite{ying2023reviving}. They use charts of SVG format to extract and model corresponding structural information between graphical elements without high computation cost.
\end{itemize}

\stitle{Method}.
\begin{itemize}
    \item \emph{Graph Nerual Network for structured dynamic charts}. Ling \etal~\cite{ying2023reviving} presented an automated method that transforms static charts into dynamic live charts for more effective communication and expressive presentation.
To overcome the difficulty of generating dynamic live charts from static vector-based SVG, this study proposes using GNN for understanding chart and recovering data and visual encodings. 
Specifically, they first transform raw SVG into graph by a graph construction algorithm which extracts 5-dimensional node features including element type, node color, fill color, stroke color and stroke width; then it builds two types of edges including stroke-wise edges and element-wise edges.
The constructed graph is then fed into two GNN-based encoder, each designed for one type of edges, to generate graph representations, which are subsequently passed to multi-layer perceptron to classify each graph element.
\end{itemize}

\subsubsection{Sequence Generation}
\stitle{Data}. To generate a complete story, most studies generate a sequence of data facts and ensemble them into a complete data story.

\begin{itemize}
    \item \emph{Relational data}. As the most basic format of visualization data, relational data is also a popular input of automatic story generation. GenAI are applied to generate textual descriptions for data tables~\cite{ying2023reviving} and construct the links between visuals and narrations through data table and word inputs~\cite{shen2023data}.
    \item \emph{Time Series Data}. 
    Time series data is a widely used data type for a variety of visual analysis tasks, ranging from visual question answering to free exploration. The traditional tools are mostly designed for single-step guidance while GenAIs provide opportunities to build a continuous exploratory visual analysis process by extracting coherent data insights.
\end{itemize}

\stitle{Method}.
\begin{itemize}
    \item \emph{Large Language Model}. Ling \etal~\cite{ying2023reviving}'s work also leveraged large language models to create animated visuals and audio narrations, including narration with contextual information, narration with insights and narration rephrasing.
To further enhance the interplay between visual animation and narration in data videos, Data Player \cite{shen2023data} applies large language models to establish semantic connections between text and visualization and then recommends suitable animation presets with domain-knowledge constraints.
Specifically, the authors design special prompt engineering with few-shot
pre-defined examples illustrating how to output semantic links sequence provided the input of both data table and narration word.

\item \emph{Reinforcement learning}. Moreover, some researchers adopt RL-based sequence generation~\cite{shi2020calliope, shi2023support, wu2023socrates}.
Shi \etal~\cite{shi2023support} build a reinforcement learning-based system to support the exploratory visual analysis of time series data. It constructs the agent's state and action space with domain knowledge to generate coherent data insights sequences as visual analysis recommendations.
Specifically, the authors use the markovian decision process (MDP) model to formulate an EVA sequence as a sequence of state-action pairs.
Then, the RL-based method seeks to maximize the cumulative reward, which combines familiarity reward and curiosity reward.
In calculation of the curiosity reward, the casualCNN model is used to embedding time sequences of different length into equal lengths.
\end{itemize}

\section{Interaction}
\label{sec:interact}

GenAI methods such as large language models have demonstrated great potential for enhancing interaction in the broader field of human computer interaction~\cite{wang2024virtuwander, hou2024c2ideas}.
With GenAI methods, users can engage with the visualization charts, extracting novel insights and findings via a natural language interface, as known as Chart Question Answering.
Given a collection of well-designed visualization charts, users can effortlessly navigate through this corpus to locate their desired chart using similarity search, which has also recently incorporated some GenAI techniques.

\subsection{Visualization Retrieval}

Having established a set of well-crafted visualization charts and share online, the subsequent question that arises is how we can help users in searching for their desired visualizations within a given repository effectively and efficiently. This task is referred to as visualization retrieval. Engaging in visualization retrieval can offer significant advantages to several downstream tasks such as learning visualization design~\cite{9472937, saleh2015learning}, visualization reuse~\cite{li2022structure}, visualization corpus construction~\cite{9984953, luo2023line}, web mining~\cite{fan2022annotating, DBLP:journals/pami/DavilaSDKG21,vqs_study}, and computational journalism~\cite{DBLP:conf/cidr/CohenLYY11}.

Recently, some researchers adopt GenAI method to facilitate visualization retrieval tailored to user intent about visual structure or other features.

\subsubsection{Spatial Generation}
\stitle{Data}. Spatial generation methods for retrieval mainly involve raster visualization images.

\begin{itemize}
\item 
\emph{Visualization images}. Recently, some studies leverage GenAI for enhanced representations of raster visualization images in retrieval, such as WYTIWYR~\cite{xiao2023wytiwyr} and LineNet~\cite{luo2023line}. 
\end{itemize}

\stitle{Method}. Methods for this task mainly include Triplet autoencoder and contrastive learning.
\begin{itemize}
\item
\emph{Triplet autoencoder}. LineNet~\cite{luo2023line} addresses the problem of line chart retrieval by considering both image-level and data-level similarity.
For this purpose, a Triplet autoencoder is constructed with the backbone architecture of vision transformer~\cite{liu2021swin}.
Additionally, Luo et al.~\cite{luo2023line} also contribute a large-scale line chart corpus, named LineBench. This corpus contains over 115,000 line charts along with corresponding metadata from four real-world datasets, facilitating the study of similarity search in line chart visualizations.
\item 
\emph{Contrastive learning}.
WYTIWYR~\cite{xiao2023wytiwyr} uses a contrastive language image pretraining (CLIP)~\cite{radford2021learning} to facilitate zero-shot user intent alignment with visualization images.
\end{itemize}

\subsubsection{Graph Generation}

\stitle{Data}. The data mainly involves SVG format visualizations.
\begin{itemize}
\item
\emph{SVG}. Graph representation is also used by a recent study~\cite{li2022structure} to incorporate structural information in SVG-format visualization retrieval.
\end{itemize}

\stitle{Method}. The method mainly includes graph contrastive learning.
\begin{itemize}
\item
\emph{Graph contrastive learning}. Specifically, the InfoGraph~\cite{sun2019infograph} architecture is used, which is a contrastive graph representation learning model.
Specifically, the model adopts GNN encoder and generate embedding vector for input graph structure, where the optimization scheme takes as input pairs of graphs and maximize the mutual information between one graph and its subgraph while minimizing mutual information with the subgraph in the other graph. 
Combining the graph representations with image-level visual representations, the visualization retrieval results prove to be more structurally consistent.
\end{itemize}

\subsubsection{Discussion}
\textbf{Retrieval augmented generation (RAG)}. In the field of GenAI, a recently popular topic is how to integrate retrieval into the generation pipeline to achieve knowledge-grounded generation and reduce the uncertainty of purely blackbox GenAI models~\cite{lewis2020retrieval, liu2023reta}.
Recently some researchers also contemplate introducing such framework to visualization generation~\cite{song2022rgvisnet} to reduce task complexity and increase reliability of generated results.
This work focuses on generating DV query sequence.
However, the RAG framework has the potential to benefit many more different GenAI applications in visualization.
For example, for infographics generation, we can take the best of both worlds by combining the previous retrieval-based methods~\cite{shi2022supporting, coelho2020infomages} with the latest purely GenAI methods~\cite{xiao2023let}. 
In this way, users can benefit from both the reliable real-world examples and the creativity of GenAI models.

\textbf{Multi-modal composed retrieval}. WYTIWYR~\cite{xiao2023wytiwyr} introduces a retrieval prototype with the novel composed query which combines image input with text describing users' additional intent.
Such multi-modal composed retrieval has attracted considerable attention in the general domain of image retrieval~\cite{baldrati2022effective, ye2023contemporary,zeng2024intenttuner} and holds promise for improving the retrieval interaction for user intent alignment.
Future work can further investigate the different combinations of multi-modal queries beyond text and image modalities and different logical compositions to allow for more flexible query interaction.

\subsection{Chart Question Answering}
In data analysis with information visualization, sometimes only providing the charts to the users is not adequate as it might be time-consuming for them to comprehend complex information about the data in the chart. 
Chart question answering (CQA)~\cite{kahou2017figureqa, zou2020affinity, kim2020answering, masry2023unichart, song_2023_gvqa, hoque_2022_vqa} is a burgeoning field of research which seeks to develop intelligent algorithms and systems to answer users' questions about the charts to expedite data analysis and enhance user interaction.

\subsubsection{Sequence Generation}

\stitle{Data}. CQA mainly considers chart questions data.
\begin{itemize}
\item
\emph{Chart questions}.
Chart questions can be categorized according to different attributes~\cite{hoque_2022_vqa}, including factual/open-ended, visual/non-visual and simple/compositional.
In addition, in a more general sense, there can be different modalities of input as query about the chart.
\end{itemize}

\stitle{Method}. The CQA methods mainly include chart elements detection and vision-language model.
\begin{itemize}
\item
\emph{Chart elements detection}.
Many AI-powered CQA models rely on detection of chart elements and structure to facilitate extraction of relevant information from visualization as explicit prior for generation of answers, such as 
PlotQA~\cite{methani2020plotqa}, FigureNet~\cite{reddy2019figurenet}, DVQA~\cite{kafle2018dvqa}, STL-CQA~\cite{singh2020stl} and LEAF-QA~\cite{chaudhry2020leaf}.
For example, PlotQA~\cite{methani2020plotqa} utilizes both Visual Element Detection (VED) and Object Character Recognition (OCR) to extract key information from charts.
In visual element detection, Faster R-CNN is used while in the OCR a traditional method is adopted.
Subsequently, the extracted chart elements is converted into a knowledge graph, which is combined with log-linear ranking of logical forms extracted from the question with  compositional semantic parsing to generate the answer.
Other works adopt more complex model fully based on neural network.
For example, DVQA~\cite{kafle2018dvqa} develops a multi-output model that is capable of answering both generic questions and chart specific questions.
The model contains an OCR sub-network composed of a CNN-based bounding box predictor and a GRU-based character-level decoder to extract the text.
Based on the results of OCR sub-network, DVQA improves the Stacked Attention Network (SAN)~\cite{yang2016stacked} for general visual question answering with additional dynamic encoding, which can adapt to chart specific vocabulary.
Although some most recent works still~\cite{masry2022chartqa} rely on Chart elements detection and data extraction like ChartOCR~\cite{luo2021chartocr} as an integral part, a few study~\cite{masry2023unichart} start to utilize a slightly different type of algorithm, which is OCR-free document image understanding such as Donut~\cite{kim2022donut}.
Donut adopts the Swin Transformer architecture and is pretrained on an OCR-pseudo task.
However, in the inference stage it does not require external explicit OCR information and simply acts like an image encoder.

\begin{figure}
    \centering
    \includegraphics[width=0.8\textwidth]{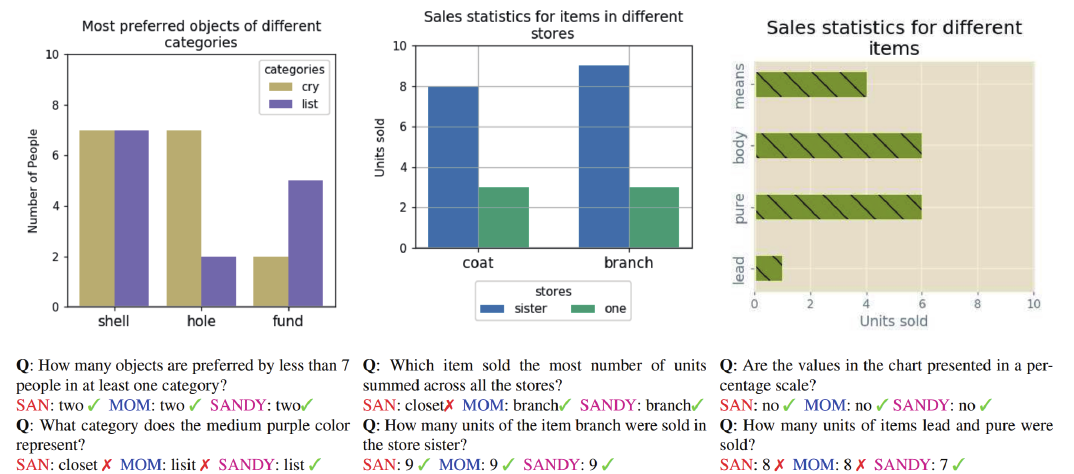}
    \caption{\revise{Chart question answering examples in DVQA~\cite{kafle2018dvqa}.}}
    \label{fig:CQA}
\end{figure}

\item
\emph{Vision-language model for multi-modal fusion}.
With the growing power of generative AI, especially the multi-modal feature fusion capability, some recent studies simplifies the chart question answering pipeline with unified vision-language model, such as ChartQA~\cite{masry2022chartqa}, PReFIL~\cite{kafle2020answering}, Unichart~\cite{masry2023unichart} and ChartLlama~\cite{han2023chartllama}.
For example, ChartQA~\cite{masry2022chartqa} builds a baseline model utilizing VL-T5~\cite{cho2021unifying}, a pre-trained unified vision-language model for text generation conditioned on multi-modal inputs. 
ChartQA also proposes their own model, VisionTaPas, which is a multi-modal extension of the TaPas~\cite{herzig2020tapas} model.
The original TaPas model is designed for answering questions over table, where a table is flattened into a sequence of words, converting the problem into essentially a unimodal-input text generation task.
For this task, the BERT~\cite{devlin2018bert} architecture is extended with additional embeddings to represent table structure and context, including embeddings of segment, row/column, rank, and previous answer.
In the VisionTaPas model, a Vision Transformer (ViT)~\cite{dosovitskiy2020image} model is utilized to extract chart image features into embeddings, as ViT has proven to be more powerful than CNN in many vision tasks.
Next, a cross-modality encoder is constructed to fuse the multi-modal embeddings of ViT and TaPas, combining information of both text and chart images for end-to-end generation of answers.
Some recent research also explores instruction tuning of pre-trained large vision-language model for more flexible generation of answer~\cite{li2023scigraphqa}.
\end{itemize}

\subsubsection{Discussion}
\textbf{Other modalities}.
Most existing CQA systems only consider single modality natural language input as the primary means of interaction with visualization.
However, other studies have shown the importance of interactions in other modalities, such as body movement~\cite{andrews2011information}, touch and pen~\cite{walny2012understanding} and gesture~\cite{badam2016supporting}.
A few studies have attempted to achieve multi-modal inputs by combining natural language or speech with mouse, pen or touch interactions~\cite{hoque2017applying, srinivasan2020inchorus, DBLP:conf/sigmod/TangLOLC22}.
Nevertheless, these tentative works rely on traditional rule-based methods for quick prototyping and have not exploited the latest GenAI methods for CQA as introduced above.
To achieve multi-modal CQA, data-driven generative AI promises more flexibility than traditional methods in diverse real-world scenarios.
For example, some researchers have started to utilized GPT to support sketch interaction for generation of chart findings documentation~\cite{lin2023inksight}.

\textbf{Combination with data embedding}.
As shown in our introduction of the algorithms, most GenAI-based CQA methods still depend on explicit detection of chart elements and underlying data for generation of precise answers.
Such detection may not be always accurate and robust for complex real-world visualization images due to the diverse styles and visualization types as well as additional noises.
One possible strategy to circumvent this issue is combining the data embedding method introduced in Section~\ref{ssec:data_embed} with CQA.
With additional information about the data embedded in the chart images, the performance of CQA can be expected to further improve.

\textbf{The promise of more precise vision-language model}.
Recent development of vision-language model is showing a trend of higher precision for more fine-grained detail in the images.
Segment Anything~\cite{kirillov2023segment} Model can locate specific semantic segment in the image given users visual or textual prompt.
Similarly, Grounding-Dino~\cite{liu2023grounding} can even more accurately generate bounding box for particular objects in images with users' prompts.
In addition, LlaVa~\cite{liu2023visual} allows users to flexibly ask about different levels of image contents from overall features to details.
For the task of CQA which requires so much precision that additional detection models are needed, these powerful vision-language models have the potential to significantly simplify the pipeline, leading towards a universal multi-modal model for chart interaction. 

\revise{
\textbf{GenAI for visual analytics}. Going beyond the interaction with single charts, some researchers recently are exploring the possibility of extending GenAI, particularly large language models to more complicated visual analytics workflow~\cite{zhao2024leva, liu2023ava}.
For example, LEVA~\cite{zhao2024leva} utilizes LLM to assist multiple stages of visual analytics including onboarding, exploration, and
summarization. 
With the development of LLM agent technology~\cite{lu2024agentlens}, GenAI can potentially take the role of humans in some visual analytics tasks.
It is an important question for future research to define new paradigm for human-AI collaboration in data visualization and analysis.
}

\section{Research Challenges And Opportunities}
\label{sec:challenge}

\subsection{Evaluating GenAI for Visualization}
The increasing use of GenAI in the production of complex and creative visualizations.
Given the essential role of rigorous evaluation in visualization design, it becomes crucial to apply similar assessment standards to AI-generated visualizations.
The distinct characteristics and challenges presented by AI-driven visualization processes necessitate careful adaptation of evaluation metrics and methodologies.
While traditional metrics such as efficiency~\cite{tufte2001visual} and aesthetic~\cite{harrison2015infographic} remain fundamental in evaluating AI-generated visualizations, the advent of AI techniques introduce additional, specific metrics that must be considered.
From the migration of assessment metrics for GenAI, the following assessment metrics are likely to be considered for evaluating differnet application of GenAI in visualization.

\begin{itemize}

\item
\textbf{Accuracy and Fidelity.} 
Ensuring accuracy and fidelity in AI-generated visualizations is paramount, particularly when applying stylization techniques. 
Techniques such as semantic contextualization in visualization \cite{xiao2023let} face the challenge of balancing data integrity with aesthetic appeal. 
This is crucial because real-life objects often do not conform to the rigid outlines typical in model-generated images, posing a risk to the accuracy of the visual representation.

\item
\textbf{Intent Alignment and Controllbility.} 
This criterion assesses the degree to which AI-generated visualizations align with the user's intent and their ability to influence the outcome. 
In Natural Language Interaction (NLI), controllability pertains to the user's efficacy in steering the output of language-based AI systems during iterative interactions \cite{maddigan2023chat2vis, yen2023coladder}. 
Furthermore, in end-to-end generation processes, it is essential that the AI-generated visualizations are sensitive to and align with the user’s specific requirements, such as style or query objectives.

\item
\textbf{Robustness and Consistency.} 
Lastly, evaluating the robustness and consistency of AI-generated visualizations across different scenarios is a key metric, ensuring reliability and applicability in diverse contexts~\cite{agrawal2023reassessing}.
Regarding LLM, it may blending fact with fiction and generating non-factual content, which called hallucination problem~\cite{ji2023survey}.
For example, in doing vqa tasks, especially in domains with specific requirements for accuracy, evaluating the \revise{hallucinations} of the generated content is essential.

\item
\textbf{Bias and Ethics.} The potential biases inherent in AI algorithms and the ethical implications of their outputs necessitate careful examination.The generative model may face potential criticism on copyright or bias issues, as the training process digests a huge amount of data obtained from the web, which is unfiltered and imbalanced.

\end{itemize}
In short, the field needs to update evaluation methods and criteria continually to keep pace with advancing GenAI technologies in assessing AI-generated visualizations.

\begin{table}[t]
\center
\scriptsize
\caption{Examples of GenAI4VIS datasets.}
\begin{tabular}{|c|c|c|c|}\hline
\textbf{Dataset} & \textbf{Data Format}  & \textbf{Source} & \textbf{Supported Tasks} \\
\hline
VizNet~\cite{hu2019viznet} & Real world tables  & Web-crawled & Data inference\\
\hline
VIS30K~\cite{chen2021vis30k} & Chart images  & Extracted from papers & Data embedding\\
\hline
Data2Vis~\cite{dibia2019data2vis} & Table-code pairs  & Synthetic & Table2VIS generation \\
\hline
nvBench~\cite{nvbench} & NL-code pairs  & Synthetic & NL2VIS generation \\
\hline
MV~\cite{chen2020composition} & MV-layout labels  & Extracted from papers & Layout transfer \\
\hline
Chart-to-text~\cite{kantharaj2022chart} & Chart-text pairs & Web-crawled & Text annotation \\
\hline
Beagle~\cite{battle2018beagle} & SVG-type labels  & Web-crawled & Visualization retrieval \\
\hline
LineBench~\cite{luo2023line} & Chart-data pairs &  Synthesis with annotation & Visualization retrieval \\
\hline
PlotQA~\cite{methani2020plotqa} & Chart-QA pairs & Crowd-sourcing + Synthetic  & CQA \\ 
\hline

\end{tabular}
\label{tab:dataset}
\end{table}

\subsection{Dataset}

As GenAI is data-driven, these methods heavily depends on the training data.
Indeed, most previous works applying GenAI to visualization build their own dataset or utilize the datasets created by prior works~\cite{chen2023state}.
Even in the era of large language models which are pre-trained on much larger general purpose dataset, a domain-specific visualization dataset can serve as valuable reference and knowledge base for efficient prompting and improving the reliability of GenAI results. 
The quality, quantity, and diversity of the dataset thus have a significant impact on the generative performance and the output quality, as it determines how the GenAI model perceives and understands the patterns and semantics of the generation requirement and generated content.

In this regard, several aspects warrant special attention in future research.
First, the diversity is important, as a diverse training dataset helps the AI model learn a broader range of topics, styles and other design patterns in real-world visualization.
This diversity enables the model to generate content that is more versatile and contextually appropriate in different situations.
However, many datasets used in training GenAI4VIS models are less diverse than real-world data~\cite{ye2022visatlas}, partly because most researchers collect or synthesize their training data for prototyping of their generative methods, without sufficient consideration of more complex authentic cases.
Therefore, building on existing GenAI4VIS studies, one important direction for future improvement is understanding the lack of diversity in current training datasets and supplement them accordingly.

Second, the heterogeneous data in different formats can reduce the reusability.
For instance, visualizations data are in a wide range of forms including raster images, SVG and different types of codes like Vega-Lite and Python.
This discrepancy necessitates the curation of datasets in different formats anew for different generation tasks.
This can also be a significant limiting factor for the size of the dataset because similar data in other formats cannot be utilized.
This in turn may lead to overfitting and other difficulties in training GenAI for visualization.
To address this issue, more robust visualization retargeting methods need to be developed to align different formats of data, such as translating Vega-Lite code to Python code and extracting graphical SVG structure from raster visualization images.
For example, recently some researchers have been exploring the idea of using large language models to generate various annotations for visualization datasets~\cite{ko2023natural}.

In addition, we provide examples of some existing datasets that can be applied to different GenAI4VIS tasks, as shown in Table~\ref{tab:dataset}.
We can find that some researchers seek to address the lack of GenAI4VIS dataset by either collecting and transforming real-world data or synthesizing data.
For example, due to the lack of NL2VIS benchmarks, nvBench~\cite{nvbench, luo2021nvbench} proposes utilizing the existing NL2SQL benchmark dataset which can be transformed into NL2VIS benchmark.
Furthermore, we can see that some larger scale real-world datasets such as VIS30K~\cite{chen2021vis30k} and VizNet~\cite{hu2019viznet} so far can only facilitate data enhancement tasks in GenAI4VIS because of the lack of annotations about the visual mapping process. 
As we mentioned above, how to strike a balance between synthetic methods with high scalability and real-world data collection which requires more manual effort or a complex retargeting process can be a critical issue.
LineBench~\cite{luo2023line} is a large-scale line chart visualization corpus (with 115,000 line charts) with the associated source dataset, the underlying data D for rendering, and the rendered visualization V in the form of an image, which can facilitate the study of similarity search of line chart visualizations.
In this light, the perspective of data-centric explainable AI~\cite{wang2023visual, anik2021data} is particularly relevant.
Many visual analytics studies for explainable AI seek to help users explore the models from the data perspective to gain insights about the potential biases, yet most of these works look at general-purpose AI or GenAI models.
In other words, there is not enough self-reflection studies from the visualization domain to diagnose GenAI4VIS models along with their training data when most researchers are rushing to apply GenAI to various subtasks in the visualization pipeline.

\subsection{GenAI4VIS vs. Generative Visualization}
In the area of digital art, there is some distinction between AI art and generative art (or algorithmic art), where the latter term largely refers to traditional procedural generation algorithms with rule-based or optimization-based methods such as graph grammar or genetic algorithm~\cite{galanter2016generative}.
In contrast, AI art mostly encompasses end-to-end purely deep-learning-based generation methods such as GAN, VAE or diffusion models.
However, in the field of visualization, there is little discussion about such distinction.
In many GenAI4VIS studies, researchers often introduce a hybrid approach, integrating many rule-based constraints and procedures with partially AI-powered methods.
For example, VizML~\cite{hu2019vizml} incorporates more than 800 hand-crafted features in the input, while restricting the generation output to predicting only a few basic visual structures.
In essence, visualization has been relying on grammar-based generation which explicitly prescribe the mapping from data to visual structures and views with a suite of different codes and rules such as Vega-Lite and visual design guidelines.
To some extent, this can limit the effort to fully harness the power of GenAI, mainly because the models cannot directly learn the distribution of the final rendered images conditioned on data and user input, which is ultimately what visualization presents to users.
This is vastly different from more mature GenAI applications in other areas.
For example, in spatial generation, latest GenAI technology can skip most traditional image synthesis and 3D modeling procedures and directly render the 2D images or 3D models. 
LIDA~\cite{dibia2023lida} makes an early effort towards a more integrated GenAI4VIS pipeline.
However, LIDA's pipeline is still divided into separate sequence generation for visualization code and spatial generation for visualization stylization in a linear workflow, where the two GenAI models do not share knowledge.
One problem due to this disconnection in LIDA, for example, is that the stylization stage cannot maintain the accuracy of the visual structure with respect to the data because the image-based stable diffusion model in the second stage is completely ignorant of structural information in the previous sequence generation.
Moreover, some recent studies in AI show that merging two large pretrained models using techniques like knowledge distillation can not only produce a versatile merged model but also boost the performance for downstream tasks that require knowledge from both models~\cite{wang2023sam}, which provides inspiration for potential strategies to improve integration of GenAI4VIS models. 

In fact, the gap between visualization and GenAI pipelines is not necessarily a downside, as this signifies opportunities for future research to combine the advantages while mitigating the respective disadvantages.
On the one hand, visualization researchers can think about how to directly model the mapping between data and views in the end-to-end statistical learning framework of GenAI, which can provide more effective learning and evaluation based on the final visual representations.
For example, this means that raster visualization images from real world sources can also be directly utilized as training data as long as it is annotated with user requirement text labels, without needing further complex explicit chart element extraction for retargeting.
In this way, visualization can potentially harness the true power of multi-modal GenAI based on large pretrained vision-language models, which is showing more accurate control down to the pixel-level in recent research~\cite{yuan2023osprey}.
On the other hand, the intermediate operations like those in visualization should not be discarded by the GenAI4VIS pipeline because GenAI can be more explainable and controllable if users are allowed to inspect and intervene in the key intermediate steps.
However, such intervention should not be the superficial rule-based constraints in hybrid methods.
Instead, researchers can take inspiration from GenAI research such as ControlNet~\cite{zhang2023adding} and LayoutDiffusion~\cite{zheng2023layoutdiffusion} to embed the control into the model itself.
Alternatively, researchers can develop interactive tools to support human intervention in the generation process~\cite{liu2023autotitle}.

\section{Conclusion}
The burgeoning GenAI technology is promising for applications in the visualization domain.
Because of GenAI's impressive capacity to model the transformation and design process by learning from real data, it can benefit a range of visualization tasks like data enhancement, visual mapping generation, stylization and interaction.  
Different types of GenAI methods have been applied to these tasks due to different data structures, including sequence generation, tabular generation, spatial generation and graph generation.
With the advent of latest GenAI technology like large language model and diffusion model, new opportunities emerge to revolutionize GenAI4VIS methods.
However, task-specific challenges still exist due to the unique characteristics of visualization tasks, which demands further investigation.
Moreover, general challenges in evaluation and datasets require some rethinking about the GenAI4VIS pipeline beyond simply borrowing state-of-the art GenAI methods.
We hope this survey can help researchers reflect on existing GenAI4VIS research from a technical perspective and provide some inspiration for future research opportunities, with a vision for improved integration of GenAI in visualization.



 \bibliographystyle{elsarticle-num} 
 \bibliography{cas-refs}





\end{document}